# Large Language Models versus Classical Machine Learning Performance in COVID-19 Mortality Prediction Using High-Dimensional Tabular Data

**Running Title:** Large Language Models vs Classical Machine Learning on Structured Data

Mohammadreza Ghaffarzadeh-Esfahani[1,2], Mahdi Ghaffarzadeh-Esfahani[2], Aryan Salahi-Niri[1], Hossein Toreyhi[1], Zahra Atf[3], Amirali Mohsenzadeh-Kermani[2], Mahshad Sarikhani[4], Zohreh Tajabadi[5], Fatemeh Shojaeian[6], Mohammad Hassan Bagheri[2], Aydin Feyzi[7], Mohammadamin Tarighatpayma[4], Narges Gazmeh[7], Fateme Heydari[4], Hossein Afshar[7], Amirreza Allahgholipour[7], Farid Alimardani[7], Ameneh Salehi[4], Naghmeh Asadimanesh[4], Mohammad Amin Khalafi[4], Hadis Shabanipour[7], Ali Moradi[7], Sajjad Hossein Zadeh[7], Omid Yazdani[4], Romina Esbati[4], Moozhan Maleki[7], Danial Samiei Nasr[4], Amirali Soheili[4], Hossein Majlesi [4], Saba Shahsavan [4], Alireza Soheilipour[4], Nooshin Goudarzi[1], Erfan Taherifard[8], Hamidreza Hatamabadi[9], Jamil S. Samaan[10], Thomas Savage[11], Ankit Sakhuja[12], Ali Soroush[12], Girish Nadkarni[12], Ilad Alavi Darazam[13,14*], Mohamad Amin Pourhoseingholi[*15], Seyed Amir Ahmad Safavi-Naini *[1,12]

1. Research Institute for Gastroenterology and Liver Diseases, Shahid Beheshti University of Medical Sciences, Tehran, Iran
2. Faculty of Medicine, Isfahan University of Medical Sciences, Isfahan, Iran
3. Faculty of Business and Information Technology, Ontario Tech University, Oshawa, Canada
4. School of Medicine, Shahid Beheshti University of Medical Sciences, Tehran, Iran.
5. Digestive Disease Research Institute, Tehran University of Medical Sciences, Tehran, Iran
6. Department of Surgery, The Johns Hopkins University, Baltimore, MD, USA
7. Student Research Committee, School of Nursing and Midwifery, Shahid Beheshti University of Medical Sciences, Tehran, Iran
8. MPH department, Shiraz University of Medical Sciences, Shiraz, Iran
9. Department of Emergency Medicine, School of Medicine, Safety Promotion and Injury Prevention Research Center, Imam Hossein Hospital, Shahid Beheshti University of Medical Sciences, Tehran, Iran
10. Karsh Division of Gastroenterology and Hepatology, Cedars-Sinai Medical Center, 8700 Beverly Blvd, Los Angeles, CA 90048.
11. Department of Medicine, Stanford University, Stanford, California, USA
12. Division of Data Driven and Digital Health (D3M), The Charles Bronfman Institute for Personalized Medicine, Icahn School of Medicine at Mount Sinai, New York, NY, USA
13. Infectious Diseases and Tropical Medicine Research Center, Shahid Beheshti University of Medical Sciences, Tehran, Iran
14. Department of Infectious Diseases, Loghman Hakim Hospital, Shahid Beheshti University of Medical Sciences, Tehran, Iran
15. National Institute for Health and Care Research (NIHR), Nottingham Biomedical Research Centre, Hearing Sciences, Mental Health and Clinical Neurosciences, School of Medicine, University of Nottingham, Nottingham, UK

*** Correspondence to:** Seyed Amir Ahmad Safavi-Naini (sdamirsa@ymail.com), Mohamad Amin Pourhoseingholi (aminphg@gmail.com), and Ilad Alavi Darazam (ilad13@yahoo.com)

## Abstract

This study compared the performance of classical feature-based machine learning models (CMLs) and large language models (LLMs) in predicting COVID-19 mortality using high-dimensional tabular data from 9,134 patients across four hospitals. Seven CML models, including XGBoost and random forest (RF), were evaluated alongside eight LLMs, such as GPT-4 and Mistral-7b, which performed zero-shot classification on text-converted structured data. Additionally, Mistral-7b was fine-tuned using the QLoRA approach. XGBoost and RF demonstrated superior performance among CMLs, achieving F1 scores of 0.87 and 0.83 for internal and external validation, respectively. GPT-4 led the LLM category with an F1 score of 0.43, while fine-tuning Mistral-7b significantly improved its recall from 1% to 79%, yielding a stable F1 score of 0.74 during external validation. Although LLMs showed moderate performance in zero-shot classification, fine-tuning substantially enhanced their effectiveness, potentially bridging the gap with CML models. However, CMLs still outperformed LLMs in handling high-dimensional tabular data tasks. This study highlights the potential of both CMLs and fine-tuned LLMs in medical predictive modeling, while emphasizing the current superiority of CMLs for structured data analysis.

# Introduction

The rapid advancement of large language models (LLMs) has revolutionized their practical applications across various domains, including medicine. These sophisticated models, trained on vast datasets, excel in a wide array of natural language processing tasks, demonstrating remarkable adaptability in assimilating specialized information from diverse medical fields [1]. While primarily designed for next-word prediction, LLMs have emerged as powerful, evidence-based knowledge assistants for healthcare providers, offering valuable insights and support in clinical decision-making processes [2–4]. While their main training centers on predicting the next word, LLMs can act as evidence-based knowledge helpers for healthcare providers, offering valuable insights and assistance [5].

In medical and clinical practice, machine learning models, particularly classical machine learning (CML) models (i.e., feature-based algorithms that learn patterns from preprocessed data rather than raw inputs), have gained significant traction in predicting patient outcomes, prognoses, and mortality rates. These models typically employ supervised and unsupervised learning methods, which primarily utilize structured data [6]. However, clinical datasets often present a complex interplay of structured and unstructured information, with clinical notes serving as prime examples of the latter. Traditionally, patient information management via machine learning has followed a two-step approach: transforming unstructured textual data into a structured format, followed by training CML models on these structured datasets. This process, however, often leads to potential information loss and introduces complexities in model deployment, hindering practical application in clinical settings [7].

While the efficacy of LLMs in handling unstructured text is well documented [8], their performance in handling structured data and their comparative effectiveness against CML models remain a critical area of investigation. This is particularly relevant given that much of the historical medical data are stored in structured formats that are often difficult to integrate [9]. **Table 1** summarizes previous studies comparing the performance of LLMs and CML approaches in medicine [10–14]. Studies reported varied results, primarily due to differences in evaluated tasks (number of input features, sample size, and prediction complexity) and transformation techniques (e.g., transforming tables into textual prompts). However, they focus on tasks with a limited number of features (<12), fail to represent real-world medical decisions, and train instances for the models (<1000), limiting the CMLs to reach their maximum performance.

Our study aims to address this knowledge gap by evaluating LLMs' predictive capabilities in the context of COVID-19 mortality prediction via a high-dimensional dataset and simple table-to-text transformation. By utilizing a sufficient number of training instances, we provide the opportunity for CMLs to reach their maximum performance, enabling a more robust comparison with LLMs. This investigation is designed to provide insights into CML versus LLM comparisons in real-world, time-sensitive, and complex clinical tasks.

**Table 1. Summary of studies comparing the performance of large language models and classical machine learning methods in medicine using structured data**

| First Author; Year | Aim and Task | Dataset (Sample Size; #feature) | Transformation Techniques | Model, Experiment, and Metric | Zero-shot performance | Training Size: Performance |
|---|---|---|---|---|---|---|
| Hegselmann; 2023 (TabLLM) [10] | To transform table-to-text for binary classification of coronary artery disease and diabetes | Diabetes (768; #7); Heart (918, #11) | Template; Billion parameter LLM; Million parameter LLM | | | |
| | | | | TabLLM-Diabetes (AUC) | 0.82 | 32: 0.68<br>512: 0.78 |
| | | | | XGBoost-Diabetes (AUC) | - | 32: 0.69<br>512: 0.80 |
| | | | | TabLLM-Heart (AUC) | 0.54 | 32: 0.87<br>512: 0.92 |
| | | | | XGBoost-Heart (AUC) | - | 32: 0.88<br>512: 0.92 |
| Wang; 2024 (MediTab) [11] | To evaluate MediTab (GPT3.5) on seven medical classification tasks and compare it with TabLLM and CML | Seven datasets of breast, lung, and colorectal cancer from clinical trials (average 1451 ranging from 53 to 2968; average #3 categorical, #15 binaries; #7 numerical) | BioBERT-based model fine-tuned on transformation and GPT3.5 for sanity check | | | |
| | | | | MediTab (Average AUC) | 0: 0.82 | 200: 0.84 |
| | | | | XGBoost (Average AUC) | 10: 0.64 | 200: 0.79 |
| Cui; 2024 (EHR-CoAgent) [12] | To investigate the efficacy of LLMs-based disease prediction using structured EHR data generated from clinical encounters (MIMIC: acute care condition in the next hospital visit; CRADLE: CVD in diabetic patients) | MIMIC-III (11,353; #?); CRADLE (34,404; #?) | Disease, medicatin, and procudere codes by mapping the code value to code name (+ prompt engineering techniques) | | | |
| | | | | EHR-CoAgent-GPT4 – MIMIC (Accuracy; F1) | 0.79; 0.73 | - |
| | | | | GPT-4 – MIMIC (Accuracy; F1) | ZSC: 0.51; 52%<br>Prompt engineered: 0.62; 0.58 | Few-shot (N=6): 0.65; 0.64 |
| | | | | GPT-3.5 – MIMIC (Accuracy; F1) | ZSC: 0.78; 0.68<br>Prompt engineered: 0.72; 0.42 | Few-shot (N=6): 0.76; 0.63 |
| | | | | RF – MIMIC (Accuracy; F1) | N=6: 0.69; 0.63 | 11,353: 0.78; 0.70 |
| | | | | LR – MIMIC (Accuracy; F1) | N=6: 0.48; 0.56 | 11,353: 0.79; 0.73 |
| | | | | DT – MIMIC (Accuracy; F1) | N=6: 0.71; 0.51 | 11,353: 0.81; 0.76 |
| | | | | EHR-CoAgent-GPT4 – CRADLE (Accuracy; F1) | 0.70; 0.60 | - |
| | | | | GPT-4 – CRADLE (Accuracy; F1) | ZSC: 0.21; 0.22; Prompt engineered: 0.30; 0.29 | Few-shot (N=6):0.41; 0.40 |
| | | | | GPT-3.5 – CRADLE (Accuracy; F1) | ZSC: 0.56; 0.52 Prompt engineered: 0.62; 0.54 | Few-shot (N=6): 0.40; 0.40 |
| | | | | RF – CRADLE (Accuracy; F1) | N=6: 0.66; 0.51 | 34,404: 0.80; 0.57 |
| | | | | LR – CRADLE (Accuracy; F1) | N=6: 0.54; 0.48 | 34,404: 0.80; 0.59 |
| | | | | DT – CRADLE (Accuracy; F1) | N=6: 0.31; 0.31 | 34,404: 0.80; 0.52 |
| Nazary; 2024 (XAI4LLM) [13] | To evaluate the diagnostic accuracy and risk factors, including gender bias and false negative rates using LLM. In addition, a comparison with CML approaches | Heart Disease Dataset (920; #11) | Using feature name-values or transforming to simple manual textual template | | | |
| | | | | Best XAILLM: LLM+RF (F1 | ZSC: 0.741 | |
| | | | | XGBoost (F1) | - | 920: 0.91 |
| | | | | RF (F1) | - | 920: 0.74 |

*Footnote: Cui et al.'s and Nazary et al.'s studies were preprint publications.*

# Methods

## 2.1 Ethical Consideration

The study was approved by the Institutional Review Board (IRB) of Shahid Beheshti University of Medical Sciences (IR.SBMU.RIGLD.REC.004 and IR.SBMU.RIGLD.REC.1399.058). The IRB exempted this study from informed consent. Data were pseudonymized before analysis; patients' confidentiality and data security were prioritized at all levels. The study was completed under the Helsinki Declaration (2013) guidelines and all experiments were performed in accordance with Iran Ministry of Health regulations. Informed consents were collected from all individuals or their legal guardians. During the generation of LLM predictions, using the OpenAI API and Poe Web interface, we opted out of training on OpenAI and used no training-use models in Poe to maintain the data safety of patient information.

## 2.2 Study Aim and Experimental Summary

The objective of this research is to evaluate the efficacy of CMLs in comparison to LLMs, utilizing a dataset characterized by high-dimensional tabular data. We employed a previously compiled dataset and focused our experimental efforts on the task of classifying COVID-19 mortality. As illustrated in Figure 1, the primary experiment encompasses **the following:**

- Assessment of the performance of seven CML models on both internal and external test sets
- The assessment of eight LLMs and two pretrained language models on the test set.
- Assessment of a trained LLM's performance on both internal and external tests.

Additionally, we investigate the performance of models necessitating training (CML and trained LLM) across varying sample sizes, coupled with an elucidation of model prediction mechanisms through SHAP analysis.

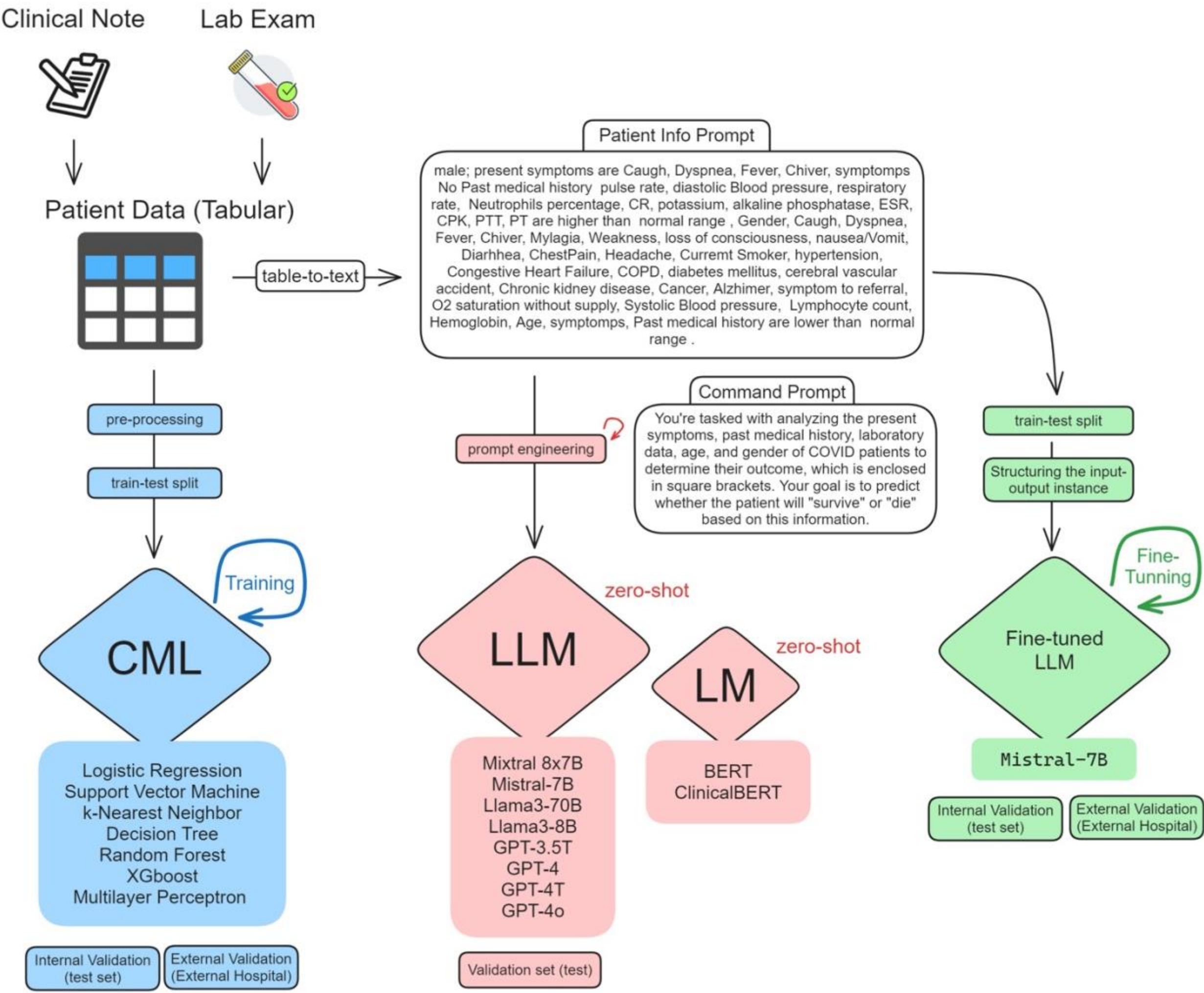

**Figure 1. Study Design and Experimental Summary**

*Image caption: This figure illustrates the design and workflow of the study, which compared large language models (LLMs) and conventional machine learning (CML) approaches for the prediction of COVID-19 patient mortality. The patient data included demographics, symptoms, past medical history, and laboratory results. The data undergo preprocessing before being structured into input–output instances. The CML pipeline involves training and validation via various algorithms, such as logistic regression, support vector machine, and random forest. Moreover, the LLM pipeline involves a prompt engineering loop. We also fine-tuned one LLM, Mistral-7b, by giving the input and ground truth. The study aims to predict patient outcomes (survival or death) on the basis of the provided information.*

## 2.3. Study Context, Data Collection, and Dataset

This study was conducted as part of the Tehran COVID-19 cohort, which included four tertiary centers with dedicated COVID-19 wards and ICUs in Tehran, Iran. The study period was from March 2020 to May 2023 and included two phases of data collection. The protocol and results of the first phase have been published previously. The four COVID-19 peaks during this period covered the alpha, beta, delta, and Omicron variants.

All admitted patients with a positive swab test during the first two days of admission or those with CT scans and clinical symptoms were included in the study. A medical team collected the patients' symptoms, comorbidities, habitual history, vital signs at admission, and treatment protocol through the hospital information system and reviewed the medical records. Laboratory values during the first and second days of admission were collected and organized from the hospitals' electronic laboratory records using pandas (v1.5.3), and NumPy (v1.24.1). Patients with a negative PCR result in the first two days of admission or with one missing clinical record in the HIS were excluded.

The dataset included the records of 9,134 patients with COVID-19. The data were filtered to include demographic information, comorbidities, vital signs, and laboratory results collected at the time of admission (first two days).

## 2.4. Computational Environment

All classical machine learning (CML) experiments were performed on a workstation equipped with an Intel Core i9-12900K CPU, 64 GB of RAM, and an NVIDIA RTX 3090 GPU (24 GB VRAM), running Ubuntu 22.04 and Python 3.10. The primary packages utilized include scikit-learn (version 1.2.2), XGBoost (version 1.7.5), pandas (version 1.5.3), and NumPy (version 1.24.1).

The fine-tuning of the Mistral-7b-Instruct model was conducted using an NVIDIA A100 80GB GPU via a cloud-based environment (Google Cloud Platform), utilizing the transformers (v4.37.2), peft (v0.9.0), and bitsandbytes (v0.41.1) libraries. The QLoRA fine-tuning procedure was implemented using 4-bit quantization, gradient accumulation steps, and mixed-precision training to optimize memory usage and reduce computational cost. All LLM zero-shot experiments were conducted via the OpenAI API and Poe interface under controlled sessions to ensure reproducibility.

## 2.5. Data Preprocessing

**Supplementary Figure S1** illustrates summary of the pipeline from raw data preprocessing through feature engineering and cleaning, to the final training–test data split used for model development and evaluation.

### 2.5.1 Imputing and normalization

The features in the dataset were divided into categorical and numerical categories. To address the missing values in the numerical features, we used an iterative imputer from the scikit-learn library. This method employs iterative prediction for each feature, considering the multiple imputation by chained equations (MICE) method (16,17). Missing values in the categorical features were imputed via KNN from the scikit-learn library. For optimal model performance, the dataset was normalized via a standard scaler (18). These preprocessing steps were executed independently for the input features of the training, test, and external validation sets, ensuring a consistent approach for handling missing values across the experimental sets without information leakage.

### 2.5.2 Feature Selection

The dataset comprised 81 on-admission features. The dataset was separated into external and internal validations using patient hospitals. Patients from Hospital-4 were used for external validation, whereas patients from the remaining hospitals were used for internal validation. For internal validation, we split the data with a test size of 20% and allocated 80% for training.

The output features in this study include "in-hospital mortality," "ICU admission," and "intubation," with a focus solely on "hospital mortality" as the targeted feature, excluding other output features. Of the 81 features initially available, 76 were employed for training, comprising 53 categorical features and the remaining numerical values. During data wrangement, two duplicate features were dropped.

We strategically employed the lasso method for feature selection because of its effectiveness in handling high-dimensional data. The Lasso method introduces regularization by adding a penalty term to the linear regression objective function, which encourages sparsity in the feature coefficients[15,16]. This approach proved to be superior to alternative methods, facilitating notable enhancements in our results. Through the application of Lasso, we derived a refined dataset that highlighted the most impactful features on the basis of their importance, aiding dimensionality reduction. We subsequently ranked and selected the top 40 features for further analyses.

#### 2.5.3 Oversampling

To address the issue of class imbalance in our dataset, we employed the synthetic minority oversampling technique (SMOTE), a widely used method in machine learning, particularly for medical diagnosis and prediction tasks [17]. By applying SMOTE, we mitigated dataset imbalances, resulting in a more robust and reliable analysis for predicting mortality. SMOTE works by creating synthetic samples for the minority class instead of simply duplicating existing samples. It selects samples from the minority class and their nearest neighbors and then generates new synthetic samples by interpolating between these samples and their neighbors. This approach not only increases the number of samples in the minority class but also introduces new data points, improving dataset diversity. In our experiments, the SMOTE technique was applied to the training set (X_train), increasing the number of samples from 6118 to 9760.

#### 2.5.4 Preparing Data for the LLM

To prepare the data for input into the LLM, we completed all the previous steps for feature selection and sampling, but normalization was not performed. As shown in Figure 1, we converted the dataset into text. We categorized the dataset features into symptoms, past medical history, age, sex, and laboratory data. For symptoms and medical history, we considered only positive data. For age, we added 'the patient's age is' before the age number. For sex, we used 'male' and 'female.' We used the normal range of laboratory data to classify the data into the normal range, higher than the normal range, and lower than the normal range. For example, if blood pressure and oxygen saturation were higher than the normal range, we used the sentence 'blood pressure and oxygen saturation are higher than the normal range.' We considered only laboratory data that were higher or lower than the normal range. The exclusion of negative features in symptoms and past medical history, or the normal range in laboratory data, is due to limitations in LLM context windows. We then concatenated the dataset into a single paragraph for each patient, indicating their medical history.

### 2.6 CML Predictive Performance

We employed five CML algorithms: logistic regression (LR), support vector machine (SVM), decision tree (DT), k-nearest neighbor (KNN), random forest (RF), multilayer perceptron neural network (MLP), and XGBoost. The hyperparameters were optimized via a grid search and cross-validation. The full details of training and hyperparameters are provided in **Supplementary Section 1.**

## 2.7 LLM Predictive Performance

We utilized open-source and proprietary LLMs to test their predictive power on clinical texts transformed from tabular data. First, we tested different prompts to determine the most efficient prompt to use, as well as the temperature (between 0.1 and 1). Full prompts are listed in Supplementary Table S1. We then sent clinical text and commands, received the unstructured output, and extracted the selected outcome, which could be either "survive" or "die." We used different sessions for each prediction, limiting the memory of the LLM to remembering previous generations.

We tested open-source, open-weight models of Mistral-7b, Mixtral 8 × 7 B, Llama3-8b, and Llama3-70b via the Poe Chat Interface. OpenAI models, including GPT-3.5T, GPT-4, GPT-4T, and GPT-4o, were utilized via the OpenAI API. We also tested the performance of two pretrained language models, BERT [18] and ClincicalBERT [19], which are fine-tuned versions of BERT on medical text. A list of all LLMs and times of use, as well as model parameters, is available in Supplementary Table S2.

### 2.7.1 Zero-Shot Classification

Zero-shot classification is an approach in prompt engineering in which the prompt is given to the model without any training. This approach is used in transfer learning, where a model used for different purposes is employed instead of fine-tuning a new model, thereby reducing the cost of training the new model. To perform zero-shot classification, we used eight different LLMs and two LMs. We provided each patient's history as input to predict whether the patient would die or survive and then stored the results.

### 2.7.2 Fine-tuning LLM

We fine-tuned one of the open-source LLMs, Mistral-7b-Instruct-v0.2, which is a GPT-like large language model with 7 billion parameters. It is trained on a mixture of publicly available and synthetic data and can be used for natural language processing (NLP) tasks. It is also a decoder-only model that is used for text-generation tasks. Fine-tuning an LLM is usually considered time-consuming and expensive; recently, several methods have been introduced to reduce costs. We implemented the QLoRA fine-tuning approach to optimize the LLM while minimizing computational resources [20].

The model was configured for 4-bit loading with double quantization, utilizing an "nf4" quantization type and torch.bfloat16 compute data type. A 16-layer model architecture with Lora attention and targeted projection modules was employed. We used the PEFT library to create a LoraConfig object with a dropout rate of 0.1 and task type 'CAUSAL_LM'. The training pipeline, established via the transformer library, consisted of 4 epochs with a per-device batch size of 1 and gradient accumulation steps of 4. We utilized the "paged_adamw_32bit" optimizer with a learning rate of 2e-4 and a weight decay of 0.001. Mixed-precision training was conducted via fp16, with a maximum gradient norm of 0.3 and a warm-up ratio of 0.03. A cosine learning rate scheduler was employed, and training progress was logged every 25 steps and reported to TensorBoard. This methodology, which combines QLoRA with the Bitsandbytes library, enables efficient enhancement of our language model while significantly reducing resource requirements, demonstrating superior performance across various instruction datasets and model scales. A more detailed description is provided in **Supplementary Section S2**.

## 2.8. CML and LLM Performance on Different Sample Sizes

To investigate the influence of training sample sizes on model performance, we conducted a series of experiments using varying sample sizes: 20, 100, 200, 400, 1000, and 6118. Multiple models were trained using these sample sizes, and their performance was evaluated on the basis of the F1 score and accuracy metrics via an internal test set. The objective of this exploration was to gain valuable insights into the correlation between the volume of training data and the accuracy of predictive models.

## 2.9. Evaluation and Cross-Validation

The accuracy of the outputs was assessed by comparing them against a ground truth that categorized outcomes as either mortality or survival. Outputs from the LLM were similarly classified. If an LLM initially produced an undefined result, the prompt was repeatedly presented up to five times to elicit a defined prediction; these instances are documented in **Supplementary Table S2**. We evaluated the models' performance via five critical metrics: specificity, recall, accuracy, precision, and F1 score. To optimize our models, we employed a grid search strategy with accuracy as the primary criterion.

We further implemented 5-fold cross-validation on the training dataset (n=6,118). The training data were randomly partitioned into five equal-sized subsets. For each fold, four subsets were used for training while the remaining subset served as a validation set. This process was repeated five times, with each subset serving as the validation set once. We calculated performance metrics (accuracy, precision, recall, specificity, F1 score, and AUC) for each fold and reported the mean and standard deviation across all five folds.

## 2.9. Statistical Analysis

Baseline characteristics were compared between patients who died and those who survived using appropriate statistical tests based on variable type and distribution. Continuous variables were analyzed using the Mann-Whitney U test (chosen over parametric alternatives due to non-normal distributions typical of clinical data) and presented as mean ± standard deviation. Categorical variables were compared using Pearson's chi-square test. The area under the receiver operating characteristic curve (AUC) was used to illustrate the predictive capacity of each model. All statistical tests were two-sided with significance set at $P<0.05$. Statistical analyses were performed using Python 3.12 with SciPy (v1.16.2).

## 2.10. Explainability

In our study, we employed SHAP (SHapley Additive exPlanations) values to examine both the total (global) and individual (granular) impacts of features on model predictions. We normalized the numerical data via a standard scaler and adopted a model-agnostic methodology. Our model-agnostic approach involved employing XGBoost as the explainer model for LLMs prediction, which was chosen for its robust performance, as demonstrated in prior research and our own findings. SHAP values provide a clear, quantitative assessment of how each feature influences individual predictions, enhancing transparency in the model's decision-making process.

For our analysis, we used the test set for each model, generated SHAP values for every prediction, and computed the mean and standard deviation of the absolute SHAP scores. We then converted SHAP scores from a range of 0 to 1 into "global impact percentages" by dividing each feature's score by the total score of all features and multiplying by 100. We calculated the average impact percentages for both CMLs and LLMs by first averaging the SHAP scores and then determining the impact percentages. To compute the standard deviation of the impact percentages, we adjusted the average standard deviation of CML/LLM via a multiplication factor derived from the ratio of the impact score to the SHAP mean. The global impact percentage represents the proportion of each feature's impact on the predicted class across the entire dataset. A violin plot visually represents the variability of each input feature's effect on the output.

# 3. Results

Our study initially included a dataset of 9,057 patients, with a mean age of 58.40 ± 19.81 years and a male–female ratio of 1.19. The overall mortality rate in this group was 25.11% (N=1818). **Table 2** shows the distribution of variables and missing data for both the survived and mortality cohorts. We utilized an internal validation test set and an external validation set comprising 2,470 and 2,248 participants, respectively, each with a mortality rate of 50%. Additionally, the validation set for zero-shot classification included 590 patients randomly selected from the internal validation test set, with a mean age of 63.85 ± 18.37 years, a male-to-female ratio of 355:255, and a mortality rate of 50% (mortality count = 295). **Table 3** details the performance metrics of all models across internal, external, and cross-validation.

**Table 3. Model results on the internal validation test set, external validation dataset and Cross validation** (The result for cross validation shows average and standard deviation across five models). * ZSC validation dataset was created using a random sample of the internal validation dataset. Abbreviations: AU-ROC, area under the receiver operating characteristic curve; ZSC: zero-shot classification; CML, classical machine learning; LLM, large language model; LR, logistic regression; SVM, support vector machine; DT, decision tree; KNN, K-nearest neighbors; RF, random forest; XGBoost, eXtreme gradient boosting; MLP, multilayer perceptron.

| | Approach | Model | Accuracy | Precision | Recall | Specificity | F1 | AUC |
|---|---|---|---|---|---|---|---|---|
| Cross validation | | | | | | | | |
| | CML | LR | 0.74 ± 0.02 | 0.91 ± 0.01 | 0.75 ± 0.03 | 0.71 ± 0.03 | 0.82 ± 0.02 | 0.73 ± 0.02 |
| | CML | SVM | 0.75 ± 0.02 | 0.91 ± 0.0 | 0.76 ± 0.02 | 0.72 ± 0.02 | 0.83 ± 0.01 | 0.74 ± 0.02 |
| | CML | DT | 0.72 ± 0.01 | 0.86 ± 0.01 | 0.77 ± 0.01 | 0.49 ± 0.03 | 0.81 ± 0.01 | 0.63 ± 0.01 |
| | CML | KNN | 0.65 ± 0.02 | 0.88 ± 0.01 | 0.64 ± 0.02 | 0.67 ± 0.03 | 0.74 ± 0.02 | 0.65 ± 0.01 |
| | CML | RF | 0.82 ± 0.01 | 0.88 ± 0.01 | 0.90 ± 0.02 | 0.50 ± 0.04 | 0.89 ± 0.01 | 0.70 ± 0.01 |
| | CML | Xgboost | 0.80 ± 0.01 | 0.88 ± 0.01 | 0.86 ± 0.02 | 0.55 ± 0.03 | 0.87 ± 0.01 | 0.71 ± 0.01 |
| | CML | MLP | 0.77 ± 0.02 | 0.88 ± 0.01 | 0.83 ± 0.02 | 0.53 ± 0.05 | 0.85 ± 0.01 | 0.68 ± 0.02 |
| Internal Validation | | | | | | | | |
| | CML | LR | 0.74 | 0.74 | 0.75 | 0.74 | 0.74 | 0.82 |
| | CML | SVM | 0.76 | 0.73 | 0.81 | 0.71 | 0.77 | 0.85 |
| | CML | DT | 0.76 | 0.76 | 0.76 | 0.76 | 0.76 | 0.76 |
| | CML | KNN | 0.69 | 0.71 | 0.66 | 0.72 | 0.68 | 0.74 |
| | CML | RF | 0.86 | 0.84 | 0.89 | 0.83 | 0.87 | 0.94 |
| | CML | Xgboost | 0.87 | 0.88 | 0.84 | 0.89 | 0.86 | 0.95 |
| | CML | MLP | 0.76 | 0.73 | 0.81 | 0.70 | 0.77 | 0.83 |
| | Fine-tuned LLM | Fine-tuned Mistral-7b | 0.72 | 0.69 | 0.79 | 0.65 | 0.74 | 0.72 |
| External Validation | | | | | | | | |
| | CML | LR | 0.73 | 0.73 | 0.74 | 0.72 | 0.73 | 0.80 |
| | CML | SVM | 0.74 | 0.73 | 0.75 | 0.72 | 0.74 | 0.81 |
| | CML | DT | 0.73 | 0.72 | 0.73 | 0.72 | 0.73 | 0.72 |
| | CML | KNN | 0.68 | 0.71 | 0.62 | 0.75 | 0.66 | 0.73 |
| | CML | RF | 0.82 | 0.79 | 0.86 | 0.77 | 0.83 | 0.91 |
| | CML | Xgboost | 0.82 | 0.85 | 0.78 | 0.86 | 0.82 | 0.92 |
| | CML | MLP | 0.71 | 0.69 | 0.76 | 0.67 | 0.72 | 0.79 |
| | Fine-tuned LLM | Fine-tuned Mistral-7b | 0.69 | 0.69 | 0.68 | 0.69 | 0.69 | 0.67 |
| ZSC Validation* | | | | | | | | |
| | LLM | Mistral-7b | 0.51 | 0.80 | 0.01 | 1.00 | 0.03 | 0.51 |
| | LLM | Mixtral-8x7b | 0.52 | 0.94 | 0.05 | 1.00 | 0.10 | 0.52 |
| | LLM | Llama-3-8b | 0.54 | 1.00 | 0.08 | 1.00 | 0.15 | 0.54 |
| | LLM | Llama-3-70b | 0.54 | 0.89 | 0.08 | 0.99 | 0.15 | 0.54 |
| | LLM | gpt-3.5-turbo | 0.50 | 0.49 | 0.17 | 0.83 | 0.25 | 0.50 |
| | LLM | gpt-4 | 0.62 | 0.84 | 0.28 | 0.95 | 0.43 | 0.62 |
| | LLM | gpt-4-turbo | 0.57 | 0.82 | 0.19 | 0.96 | 0.31 | 0.57 |
| | LLM | gpt-4o | 0.50 | 1.00 | 0.00 | 1.00 | 0.01 | 0.50 |
| | LM | BERT | 0.50 | 1.00 | 0.00 | 1.00 | 0.01 | 0.50 |
| | LM | ClinicalBERT | 0.50 | 1.00 | 0.00 | 1.00 | 0.01 | 0.50 |

**Table 2. Data are presented as mean ± standard deviation for continuous variables and n (%) for categorical variables. P-values were calculated using Mann-Whitney U test for continuous variables and chi-square test for categorical variables.** Abbreviations: BP, blood pressure; CABG, coronary artery bypass grafting; CKD, chronic kidney disease; COPD, chronic obstructive pulmonary disease; CPK, creatine phosphokinase; ESR, erythrocyte sedimentation rate; GI, gastrointestinal; ICU, intensive care unit; INR, international normalized ratio; MCV, mean corpuscular volume; PT, prothrombin time; PTT, partial thromboplastin time; WBC, white blood cell.

| Variable | Total (N=9,057) | Mortality (n=1,818) | Survived (n=7,239) | P-value |
|---|---|---|---|---|
| ***Demographics*** | | | | |
| Age (years) | 58.4 ± 19.8 | 69.9 ± 23.5 | 55.5 ± 17.6 | <0.001*** |
| Male Gender | 54.5% (4,932) | 59.3% (1,078) | 53.2% (3,854) | <0.001*** |
| ***Vital Signs*** | | | | |
| Diastolic BP (mmHg) | 75.3 ± 11.7 | 74.1 ± 14.2 | 75.6 ± 10.8 | <0.001*** |
| O2 Saturation without Support (%) | 85.1 ± 17.0 | 79.2 ± 18.1 | 86.6 ± 16.4 | <0.001*** |
| Pulse Rate (bpm) | 88.0 ± 15.5 | 90.6 ± 19.3 | 87.3 ± 14.2 | <0.001*** |
| Respiratory Rate (breaths/min) | 19.3 ± 5.1 | 20.6 ± 6.9 | 19.0 ± 4.4 | <0.001*** |
| Systolic BP (mmHg) | 120.4 ± 18.8 | 119.9 ± 23.4 | 120.6 ± 17.3 | 0.003** |
| Temperature (°C) | 37.1 ± 1.3 | 37.1 ± 1.5 | 37.1 ± 1.3 | 0.334 |
| ***Symptoms*** | | | | |
| Abdominal Pain | 5.5% (502) | 5.2% (94) | 5.6% (408) | 0.473 |
| Anorexia | 16.1% (1,456) | 15.6% (284) | 16.2% (1,172) | 0.579 |
| Chest Pain | 8.9% (810) | 6.8% (124) | 9.5% (686) | <0.001*** |
| Chills | 26.6% (2,409) | 22.0% (400) | 27.8% (2,009) | <0.001*** |
| Cough | 48.0% (4,344) | 42.6% (774) | 49.3% (3,570) | <0.001*** |
| Diarrhea | 9.3% (842) | 7.3% (132) | 9.8% (710) | <0.001*** |
| Dyspnea | 58.1% (5,260) | 63.2% (1,149) | 56.8% (4,111) | <0.001*** |
| Ear Pain | 0.1% (12) | 0.1% (1) | 0.2% (11) | 0.480 |
| Fever | 42.4% (3,843) | 38.4% (699) | 43.4% (3,144) | <0.001*** |
| Headache | 10.5% (949) | 6.0% (109) | 11.6% (840) | <0.001*** |
| Hemiparesis | 0.7% (48) | 1.1% (15) | 0.6% (33) | 0.097 |
| Hemorrhage | 0.4% (33) | 0.8% (14) | 0.3% (19) | 0.003** |
| Joint Pain | 1.0% (87) | 1.0% (19) | 0.9% (68) | 0.780 |
| Loss of Consciousness | 8.0% (726) | 22.6% (410) | 4.4% (316) | <0.001*** |
| Myalgia | 29.0% (2,628) | 19.6% (356) | 31.4% (2,272) | <0.001*** |
| Nausea/Vomiting | 20.8% (1,888) | 17.1% (311) | 21.8% (1,577) | <0.001*** |
| Olfactory Dysfunction | 1.2% (112) | 0.4% (7) | 1.5% (105) | <0.001*** |
| Rhinorrhea | 0.9% (86) | 0.9% (16) | 1.0% (70) | 0.837 |
| Seizure | 0.5% (49) | 0.8% (15) | 0.5% (34) | 0.095 |
| Skin Lesion | 0.2% (21) | 0.6% (11) | 0.1% (10) | <0.001*** |
| Sore Throat | 2.5% (224) | 1.5% (27) | 2.7% (197) | 0.003** |
| Weakness | 36.7% (3,326) | 41.9% (762) | 35.4% (2,564) | <0.001*** |
| ***Comorbidities*** | | | | |
| Alcohol Use | 0.6% (53) | 0.9% (16) | 0.5% (37) | 0.095 |
| Alzheimer's Disease | 1.9% (170) | 5.0% (90) | 1.1% (80) | <0.001*** |
| Anemia | 1.0% (94) | 1.3% (24) | 1.0% (70) | 0.231 |
| Asthma | 2.4% (221) | 2.3% (42) | 2.5% (179) | 0.752 |
| COPD | 1.6% (144) | 2.4% (43) | 1.4% (101) | 0.004** |
| Cancer | 4.9% (446) | 8.8% (160) | 4.0% (286) | <0.001*** |
| Chronic Kidney Disease | 3.7% (331) | 6.4% (116) | 3.0% (215) | <0.001*** |
| Current Smoker | 5.0% (453) | 5.9% (107) | 4.8% (346) | 0.061 |
| Diabetes Mellitus | 26.3% (2,384) | 35.3% (642) | 24.1% (1,742) | <0.001*** |
| Fatty Liver Disease | 0.7% (61) | 0.5% (9) | 0.7% (52) | 0.379 |
| GI Disorder | 1.3% (115) | 1.7% (31) | 1.2% (84) | 0.082 |
| Heart Failure | 1.6% (146) | 2.9% (53) | 1.3% (93) | <0.001*** |
| Hepatitis C | 0.1% (11) | 0.2% (4) | 0.1% (7) | 0.248 |
| Hookah Use | 0.6% (53) | 0.4% (7) | 0.6% (46) | 0.280 |
| Hyperlipidemia | 4.6% (421) | 4.4% (80) | 4.7% (341) | 0.618 |
| Hypertension | 32.0% (2,899) | 44.3% (806) | 28.9% (2,093) | <0.001*** |
| Immunocompromised | 0.2% (19) | 0.1% (2) | 0.2% (17) | 0.399 |

| Ischemic Heart Disease | 13.5% (1,224) | 21.1% (384) | 11.6% (840) | <0.001*** |
|---|---|---|---|---|
| Opium Use | 3.9% (352) | 5.5% (100) | 3.5% (252) | <0.001*** |
| Parkinson's Disease | 0.8% (76) | 2.0% (37) | 0.5% (39) | <0.001*** |
| Pregnancy | 0.6% (38) | 0.1% (2) | 0.7% (36) | 0.033* |
| Prior CABG | 3.5% (316) | 5.9% (108) | 2.9% (208) | <0.001*** |
| Prior Pneumonia | 0.4% (39) | 0.9% (17) | 0.3% (22) | <0.001*** |
| Prior Stroke | 4.5% (404) | 9.9% (180) | 3.1% (224) | <0.001*** |
| Psychiatric Disorder | 1.7% (156) | 2.5% (46) | 1.5% (110) | 0.004** |
| Rheumatoid Arthritis | 0.9% (82) | 1.3% (24) | 0.8% (58) | 0.051 |
| Seizure Disorder | 1.3% (114) | 1.4% (26) | 1.2% (88) | 0.538 |
| Thyroid Disorder | 4.5% (405) | 4.1% (75) | 4.6% (330) | 0.462 |
| Tuberculosis | 0.3% (23) | 0.2% (4) | 0.3% (19) | 1.000 |
| ***Laboratory Values*** | | | | |
| Alkaline Phosphatase (U/L) | 226.4 ± 181.4 | 264.1 ± 223.4 | 215.5 ± 165.7 | <0.001*** |
| CPK (U/L) | 333.2 ± 996.3 | 505.5 ± 1668.4 | 283.7 ± 683.7 | <0.001*** |
| Creatinine (mg/dL) | 1.9 ± 2.5 | 2.6 ± 3.2 | 1.7 ± 2.3 | <0.001*** |
| ESR (mm/hr) | 58.1 ± 1387.2 | 40.7 ± 26.8 | 62.9 ± 1567.8 | 0.027* |
| Hemoglobin (g/dL) | 12.4 ± 2.2 | 11.9 ± 2.5 | 12.5 ± 2.1 | <0.001*** |
| INR | 5.4 ± 22.1 | 7.4 ± 25.7 | 4.9 ± 20.8 | <0.001*** |
| Lymphocytes (%) | 17.6 ± 10.7 | 13.0 ± 9.8 | 18.8 ± 10.6 | <0.001*** |
| MCV (fL) | 84.7 ± 7.4 | 85.7 ± 8.0 | 84.4 ± 7.2 | <0.001*** |
| Neutrophils (%) | 77.8 ± 12.0 | 83.2 ± 10.5 | 76.4 ± 12.0 | <0.001*** |
| PT (seconds) | 19.1 ± 27.1 | 22.9 ± 32.4 | 17.9 ± 25.3 | <0.001*** |
| PTT (seconds) | 45.1 ± 61.7 | 51.8 ± 77.2 | 43.1 ± 56.0 | <0.001*** |
| Platelets ($\times 10^9$/L) | 201.9 ± 91.2 | 193.2 ± 96.2 | 204.2 ± 89.6 | <0.001*** |
| Potassium (mmol/L) | 7.0 ± 17.8 | 8.3 ± 22.0 | 6.7 ± 16.5 | <0.001*** |
| Sodium (mmol/L) | 137.3 ± 5.7 | 137.7 ± 7.3 | 137.2 ± 5.2 | 0.025* |
| WBC ($\times 10^9$/L) | 8.3 ± 7.0 | 10.5 ± 7.5 | 7.7 ± 6.8 | <0.001 |
| ***Outcomes and Treatment*** | | | | |
| Symptom to Referral Time (days) | 6.9 ± 8.7 | 6.5 ± 8.0 | 7.0 ± 8.8 | <0.001*** |
| Dialysis | 3.3% (303) | 10.1% (183) | 1.7% (120) | <0.001*** |
| ICU Admission | 16.3% (1,474) | 48.2% (876) | 8.3% (598) | <0.001*** |
| Mechanical Ventilation | 9.7% (876) | 41.7% (758) | 1.6% (118) | <0.001*** |

## 3.1 Classic Machine Learning Predictive Performance

As shown in **Figure 2**, XGBoost and RF were the top-performing models in terms of accuracy, achieving scores of 86.28% and 86.52%, respectively. These models also excelled in precision, recall, specificity, and F1 scores, all surpassing 85%. The MLP also delivered an acceptable performance, with an accuracy of 75.87%. When the models were applied to the external validation set, a slight decline in the AUC of 2–5% was observed. Supplementary Figures S2 and S3 depict the confusion matrix of the CMLs on the internal validation test set and the external validation set, respectively. SVM, KNN, and DT showed consistent performance across both validation sets, confirming their reliability in generalizing to unseen data.

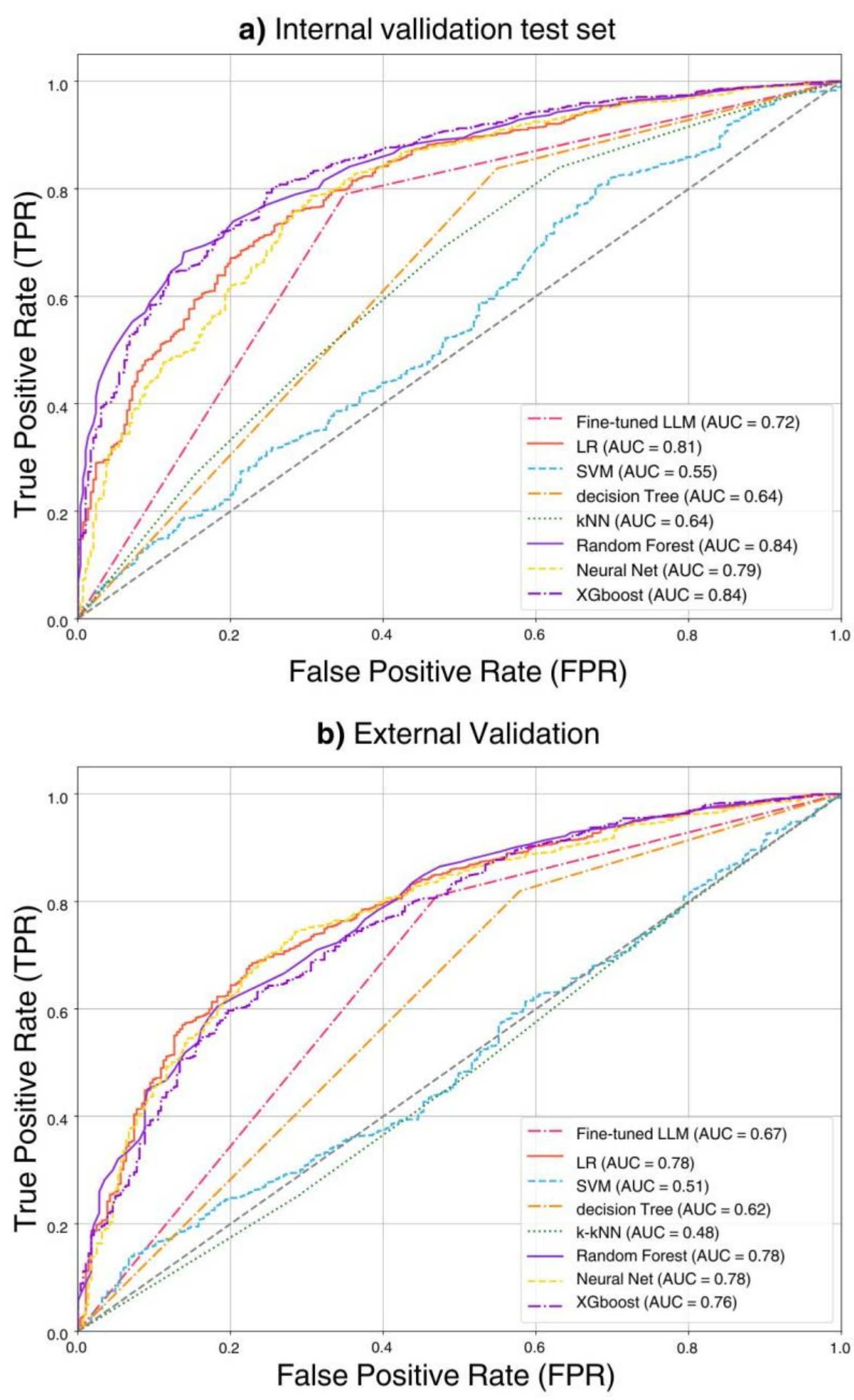

**Figure 2. ROC curves and AUC scores for COVID-19 mortality prediction models in internal and external validation.**

*Caption: Models include fine-tuned LLM (Mistral 7b) and seven classical machine learning algorithms: logistic regression (LR), support vector machine (SVM), decision tree, k-nearest neighbors (kNN), random forest, neural network, and XGBoost. Upper panel: internal validation; Lower panel: external validation. Curves show true positive rate (TPR) versus false positive rate (FPR). AUC scores indicate each model's discriminative power.*

## 3.2 LLM: Zero-shot classification and fine-tuned Mistral-7b

The zero-shot classification results showed variability among the models, with GPT-4 outperforming the other models by achieving an accuracy of 0.62 and an F1 score of 0.43 and recording the highest recall at 0.28 among the LLMs. Generally, LLMs exhibited low recall rates, predominantly classifying predictions as "mortality." The open-source models, including Llama-3-70B, Llama-3-8B, Mistral-7b-Instruct, and Mistral-8x7b-Instruct-v0.1, had F1 scores ranging from 0.03 (Mistral-7b) to 0.15 (Llama3-8B and Llama3-70b). Notably, the gpt-4o model showed limited effectiveness, with an F1 score of 0.01, indicating a challenge in distinguishing between true positives and true negatives. The pretrained language models – BERT and ClinicalBERT – also labeled all outcomes as dies, failing to provide predictive power. **Supplementary Figure S4** shows confusion matrix of LLM and language models.

Fine-tuning Mistral-7b significantly improved its performance, increasing the F1 score from 0.03 to 0.74 in the internal test set and to 0.69 in the external test set. This fine-tuned version also demonstrated a high recall rate of 78.98%, a substantial increase from 1% in zero-shot classification, showing its ability to accurately identify a greater proportion of actual survival instances. This consistency between internal and external validations highlights the generalizability of the fine-tuned Mistral-7b in mortality prediction. The confusion matrix of fine-tuned and zero-shot Mistral-7b is presented at **Supplementary Figure S5**.

## 3.3 Comparing models on different training sample sizes

To evaluate the impact of training sample size on model efficacy, experiments were conducted across various sample sizes. Figure 3 shows that the performance of all CMLs increased as the size of the training set increased. XGBoost demonstrated the strongest performance across all categories: small (100 samples), medium (400-1000 samples), and full training set sizes (6118 samples). Notably, the MLP neural network and SVM exhibited the most significant performance improvements, with accuracies increasing from 55% with 20 training samples to 73% and 77%, respectively.

**a)**

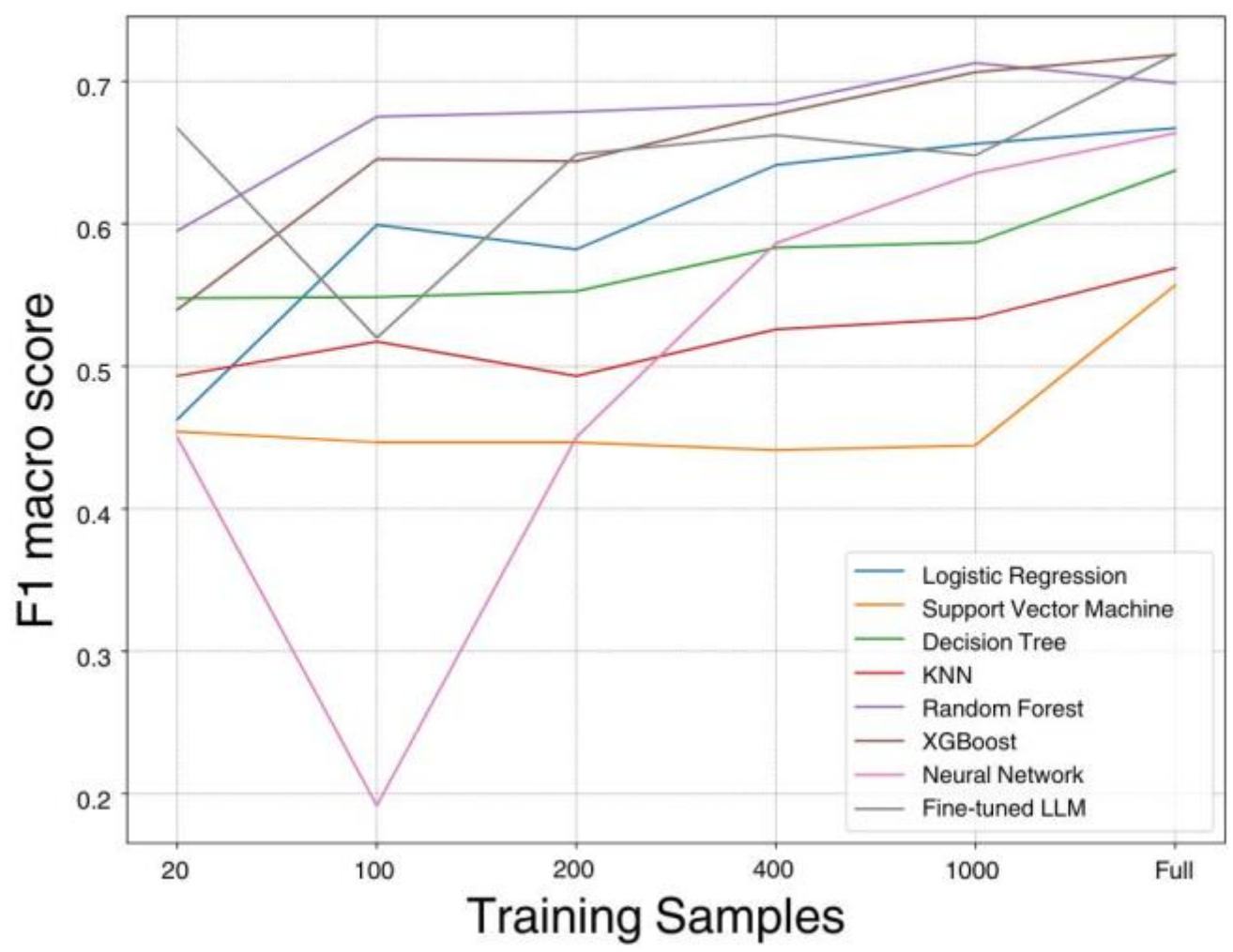

**b)**

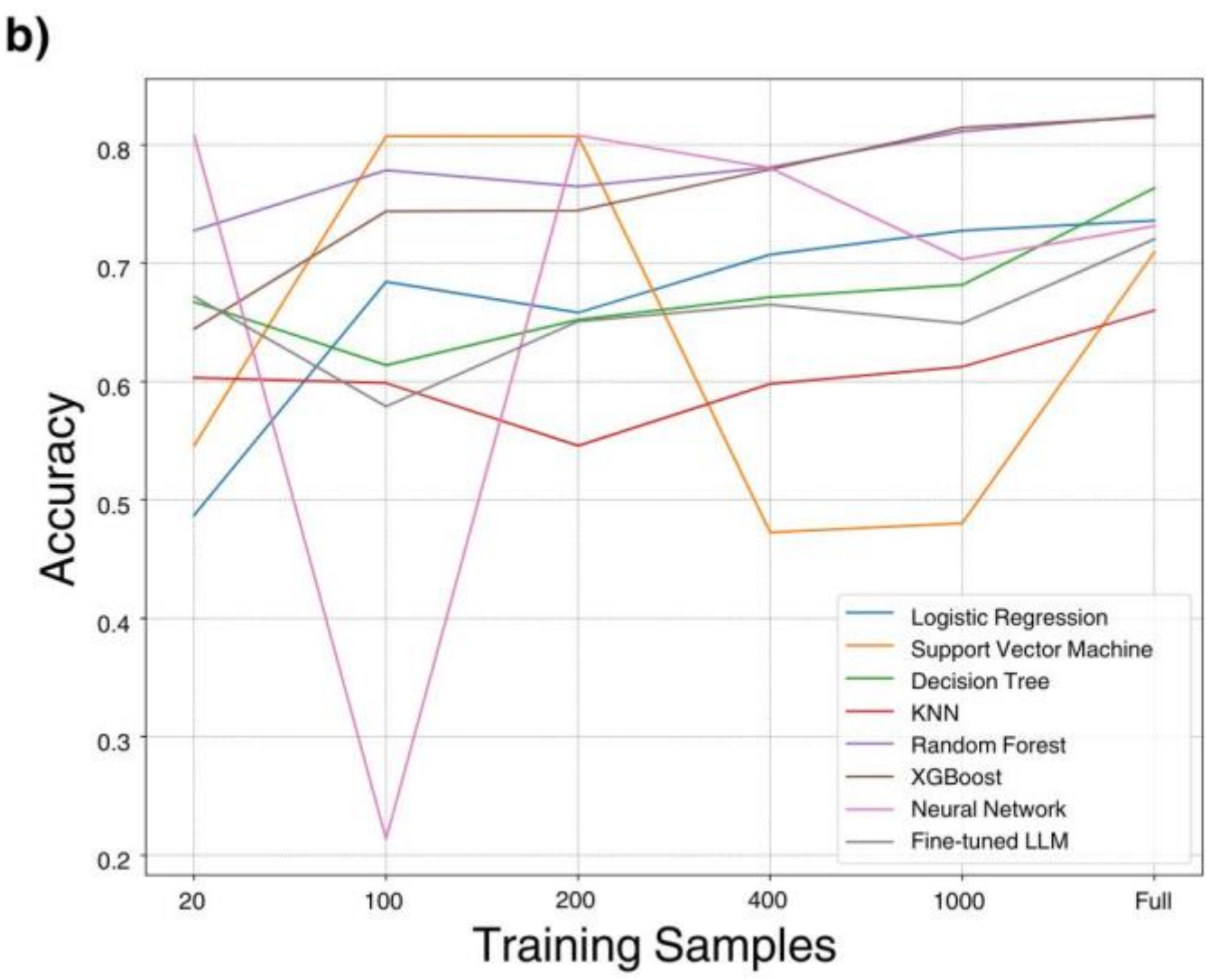

**Figure 3. Performance comparison of classical machine learning (CML) models and fine-tuned large language models (LLMs) in COVID-19 mortality prediction across varying sample sizes.**

*F1 scores and accuracy are shown for seven CML models (logistic regression, support vector machines, decision trees, k-nearest neighbors, random forests, XGBoost, and neural networks) and a fine-tuned LLM. Training sample sizes range from 20 to 2047. XGBoost consistently outperforms other models, with performance improving as sample size increases.*

In contrast, while the zero-shot performance of GPT-4 reached an F1 score of 0.43, CMLs still surpassed both zero-shot classification and fine-tuned LLMs in predicting COVID-19 mortality. During the fine-tuning of Mistral-7b, notable performance degradation occurred in scenarios with small training sizes, leading to a loss of broader model understanding, an effect termed "negative transfer."

## 3.4 Explainability: Impact of Features on Prediction

As shown in Supplementary Figure S5 while the global impact of features among CMLs exhibits similar patterns, with many of the top 10 impactful features being consistent, the granular impact differs significantly. For example, in the context of O2 saturation levels in patients, XGBoost, RF, DT, and MCP consider both high (increasing mortality risk) and low (increasing survival chance) levels to be significant, whereas KNN and LR focus only on low saturation levels. According to Figure 4.a, the most influential features are age (11.18%) and O2 saturation (9.89%), followed by LOC (4.83%), lymphocyte count (4.79%), dyspnea (3.76%), and sex (3.68%).

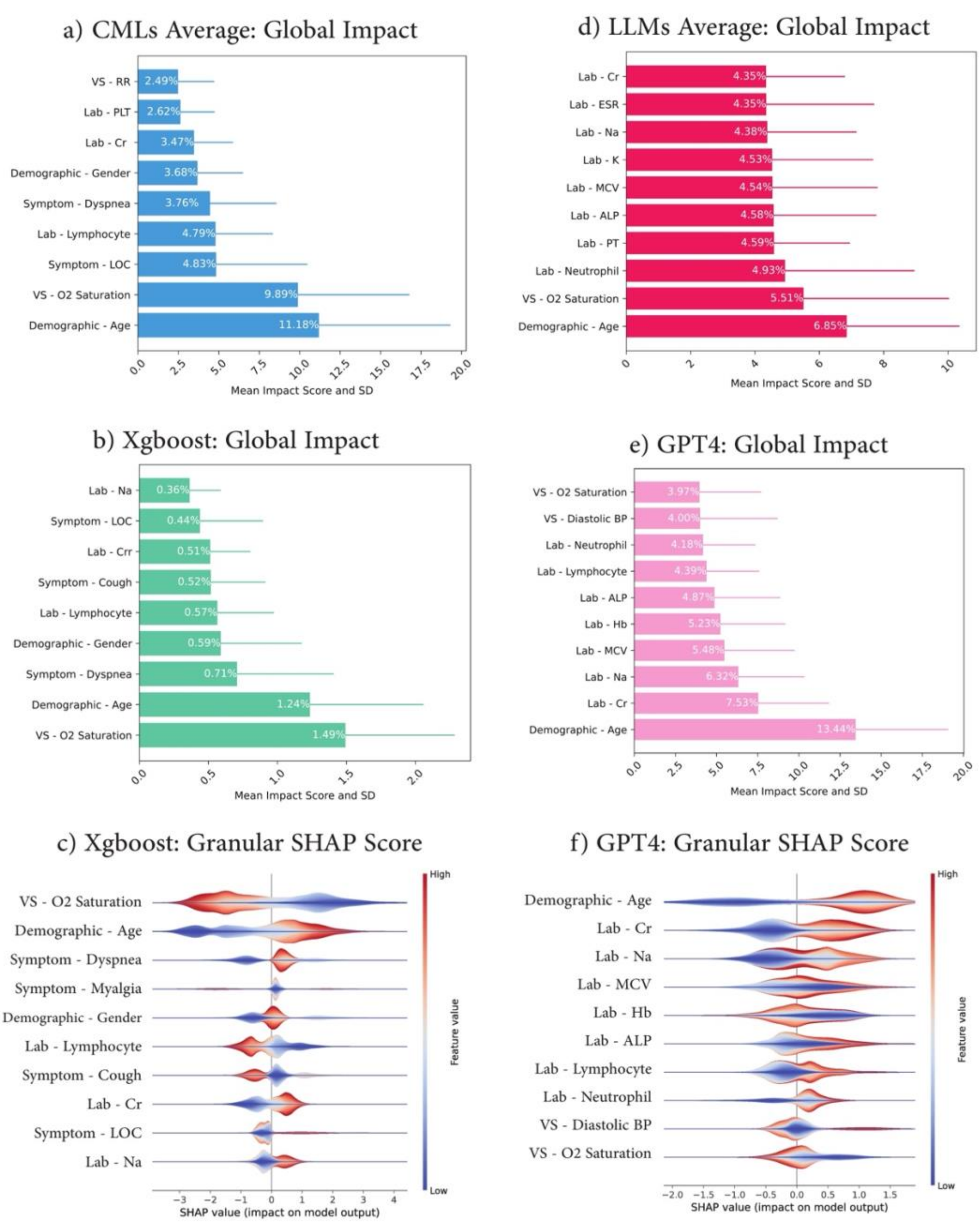

**Figure 4. SHAP analysis comparing feature importance in COVID-19 mortality prediction models.**

*(a) Average global feature impact for classical machine learning (CML) models. (b) Global impact scores for XGBoost (best-performing CML). (c) SHAP score distribution for XGBoost. (d) Average global feature impact for large language models (LLMs). (e) Global impact scores for GPT-4 (best-performing LLM). (f) SHAP score distribution for GPT-4. Key features include age, O2 saturation (VS - O2 Sat), and creatinine (Cr) levels. In panels c and f, red indicates higher feature values; positive SHAP values increase mortality prediction, negative values decrease it. VS: vital sign.*

Conversely, the influence of features in LLMs, particularly in lower-performing models such as Mistralb-7b and GPT4o, appears less coherent, as illustrated in Supplementary Figures S6 and S7. This inconsistency contributes to noise in the average feature impact among LLMs (Figure 4.d). Nonetheless, age (6.58%) and O2 saturation (5.51%) remained the most significant features, with a series of laboratory tests, including neutrophil count, PT, ALP, MCV, K, Na, ESR, and Cr, revealing impacts in the 4%-5% range.

When comparing the top performers among CMLs and LLMs—XGBoost and GPT4—the patterns of global (Fig. 4.b and Fig. 4.e) and granular (Fig. 4.c and Fig. 4.f) impacts diverge, with XGBoost displaying more specific impacts and GPT4 showing broader ranges of impact.

Figure 5 illustrates how fine-tuning Mistral-7b altered the impact of features at both the global and granular levels. This refinement in prediction logic aligned the top 10 most important features more closely with those of CMLs, resulting in more equitable impact percentages among features and enhanced granularity.

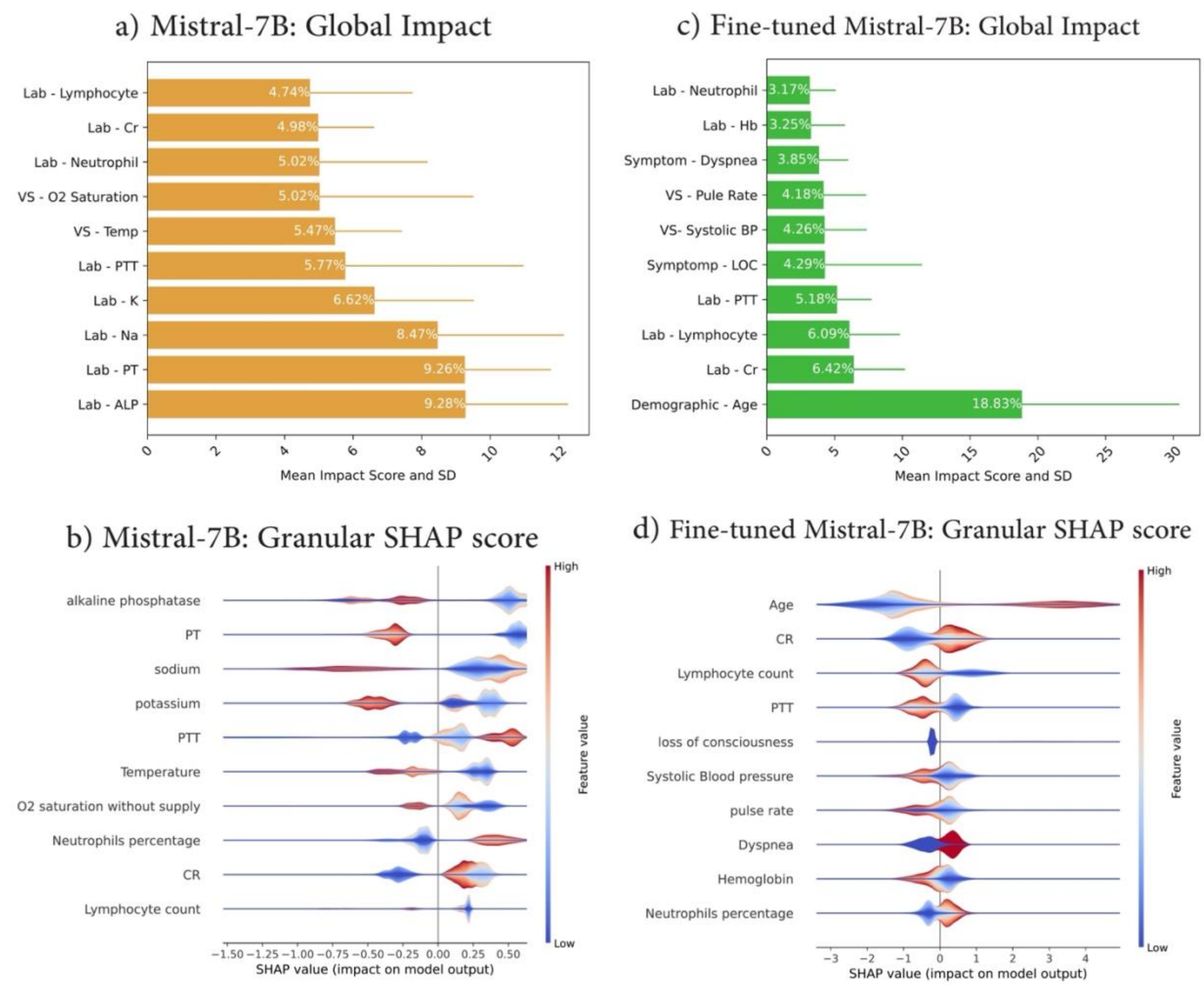

**Figure 5. SHAP analysis for Mistral-7b and fine-tuned Mistral-7b models in COVID-19 mortality prediction.**

*(a) Global feature impact for base Mistral-7b, with alkaline phosphatase (ALP), prothrombin time (PT), and sodium (Na) as top features. (b) SHAP score distribution for base Mistral-7b, showing individual feature value influences. (c) Global feature impact for fine-tuned Mistral-7b, highlighting age and creatinine (Cr) as most influential. (d) SHAP score distribution for fine-tuned Mistral-7b, emphasizing the impact of age, Cr, and lymphocyte count. In panels b and d, higher SHAP values indicate increased mortality prediction.*

### 3.5 Pipeline Validation

Supplementary Table S3 presents the XGBoost model's F1 scores for external validation, showing a result of 0.82 (AUC: 0.92) with imputation and 0.89 (AUC: 0.60) without imputation. Supplementary Table S4 presents data on external and internal validation using SMOTE to address class imbalance in CMLs. Application of SMOTE resulted in increased performance metrics for both validation sets across CMLs; for example, the XGBoost AUC rose from 0.60 to 0.92 in external validation.

## Discussion

Our study reveals a notable performance gap between CML models and LLMs in predicting patient mortality via tabular data. RF and XGBoost emerged as the top CML performers, achieving over 80% accuracy and an F1 score of 0.86. In contrast, the best-performing LLM, GPT-4, achieved 62% accuracy and an F1 score of 0.43 in zero-shot classification. This disparity highlights the challenges LLMs face when dealing with purely tabular data. Notably, increasing our sample size from 5,000 patients in our previous study to 9,000 patients in this study significantly improved the performance of CML models. The AUC of RF improved from 0.82 to 0.94, underscoring the importance of large and diverse datasets in realizing the full potential of CMLs in medical tasks.

LLM performance heavily relies on the knowledge embedded within model weights, the complexity of input data, and the table-to-text transformation technique. Our approach, which uses a simple prompt and transformation to resonate with current clinical use, achieved results comparable to those of similar studies, with F1 scores of 0.50–0.60 across different medical tasks using LLMs such as the GPT-4 or GPT-3.5 [10,11,21]. However, in line with many previous studies, we found that CMLs can outperform this zero-shot performance with even fewer than 100 training samples [10,21].

Given the performance gap between CMLs and LLMs, researchers have explored two main approaches for improving LLM performance: pipeline improvements and fine-tuning. Previous studies have shown that LLMs can close the gap in CML performance via pipeline improvements such as prompt engineering techniques (XAI4LLM), few-shot approaches (XAI4LLM, EHR-CoAgent, TabLLM), multiple runs of LLM to double-check results (EHR-CoAgent), the addition of a tree-based explainer alongside the LLM (XAILLM), or novel LLM-based text-to-table transformation (Medi-TAB, TabLLM) [22–24]. However, many of their evaluated tasks may not resonate with real-world use, as they have low-dimensional datasets (8-15 features) that do not reflect real-world complex medical data and limited sample sizes (<500 instances in rare classes) that restrict CMLs from reaching their maximum performance.

The alternative approach, fine-tuning or in-context learning, aims to modify the model weights to teach the model a new task, which has been evaluated on name entity recognition and text extraction [25,26]. We validated this approach in our high-dimensional task, where fine-tuning Mistral increased the F1 score from 0.03 to 0.69, even with a resource-efficient QLoRa method. Our SHAP analysis provides initial evidence of an improved rationale after fine-tuning, as the top 10 features more closely align with XGBoost and clinician decision-making.

Despite these advancements, LLMs still face significant limitations that affect their applicability in medical settings. Their vulnerability to hallucination raises concerns about producing harmful information [27], whereas computational constraints impose token limits that can truncate responses and diminish interaction quality. [27,28]. Data privacy is another crucial concern, particularly in medical contexts, as many powerful LLMs are proprietary or require cloud-based computations, increasing the risk of data leaks [29]. Moreover, the cost of using LLM APIs for large clinical databases can disproportionately impact low- and middle-income communities [30]. While open-source models present a more affordable alternative, they may not match the capabilities of proprietary models.

In light of these challenges, alternative approaches have emerged, including the use of small pretrained language models and rule-based systems. These offer resource-efficient alternatives to large LLMs. Previous studies have shown that rule-based and gradient boosting algorithms can achieve strong overall performance in specific tasks, such as extracting physical rehabilitation exercise information from clinical notes [31,32]. Additionally, fine-tuning pretrained BERT-like models has yielded promising results in some medical applications. However, our brief experiment with the zero-shot performance of pretrained models (BERT and ClinicalBERT) revealed their limitations, suggesting that further research is needed to optimize these approaches for complex medical tasks.

It is important to acknowledge several limitations of our study. Although the fine-tuning method used was resource efficient, it may not have been the most effective for achieving maximum performance. Fine-tuning for conversational responses instead of classification tasks with models similar to BERT may result in less reliable predictions; however, this approach mirrors how clinicians interact with AI tools. Future research could investigate ways to balance conversational accessibility and prediction accuracy. Our fine-tuned model was a small LLM with the lowest performance among our eight tested LLMs, indicating that fine-tuning larger and more accurate models could yield better results. Furthermore, our table-to-text transformation and prompts were designed to resonate with a medical user context, but more robust approaches (e.g., few-shot learning, advanced prompt engineering, and sophisticated transformation techniques) may achieve higher accuracies, especially in zero-shot classification [11,21]. Although our sample size was substantial, the retrospective nature of our investigation necessitates prospective validation to confirm the generalizability of these findings. As all participating hospitals operated within our specific resource context, variations in healthcare access and quality may have influenced the generalizability of the models to other countries and settings.

Our findings highlight several critical areas for future research in the application of LLMs to medical data analysis. We propose the following research questions to advance the field:

- Does the LLM explanation of the prediction (death or survival) in human language align with the feature importance analysis? Can LLMs accurately explain their rationale?
- What would be the performance of fine-tuning pretrained models and large LLMs compared to small LLMs?
- Could we create a model to distinguish correct answers from incorrect answers via LLM output? How can we measure the certainty of the given answer?

# Conclusion

The efficacy of LLMs versus CML approaches in medical tasks appears to be contingent upon data dimensionality and data availability. In low-dimensional scenarios with limited samples, LLM-based methodologies may offer superior performance; however, as dimensionality increases and diverse sample sizes become available, CML techniques tend to outperform the zero-shot capabilities of LLMs. Notably, fine-tuning LLMs can substantially enhance their pattern recognition and logical processing, potentially achieving performance levels comparable to those of CMLs. The potential of LLMs to process both structured and unstructured data may outweigh marginally lower performance metrics than CMLs do. Ultimately, the choice between LLMs and CMLs should be guided by careful consideration of task complexity, data characteristics, and clinical context demands, with further research warranted to elucidate the precise conditions under which each methodology excels.

## Acknowledgments

We used ChatGPT with the following prompt: "Is this paragraph grammatically correct, and can you make it sound scientific? Improve the grammar to improve the English style, understanding, and coherence". Two authors, SAASN and MG, reviewed the suggestions and accepted relevant changes. All the authors are responsible for the validity of the final draft.

## Authors' contributions

MoGE: Conceptualization, Methodology, Programming, Investigation, Writing Original Draft; MaGE: Investigation, Methodology; ASN: Investigation, Methodology; HT: Writing Original Draft, Methodology; ZA: Investigation, Programming; AMK: Investigation; MS: Investigation; ZT: Investigation; FS: Investigation; MHB: Investigation; AF: Investigation; MT: Investigation; NG: Investigation; FH: Investigation; HA: Investigation; AA: Investigation; FA: Investigation; AS: Investigation; NA: Investigation; MAK: Investigation, Project Administration; HS: Investigation; AM: Investigation; SHZ: Investigation; OY: Investigation; RE: Investigation; MM: Investigation; DSN: Investigation; ALS: Investigation; HM: Investigation; SS: Investigation; ARS: Investigation; NG: Investigation; ET: Investigation, Validation; HH: Investigation; JSS: Reviewing and Editing the Manuscript; TS: Reviewing and Editing the Manuscript; AKS: Reviewing and Editing the Manuscript; ALS: Methodology, Validation, Reviewing and Editing the Manuscript; GN: Reviewing and Editing the Manuscript; IAD: Data Acquisition, Reviewing and Editing the Manuscript, Administration, Supervision; MAP: Data Acquisition, Reviewing and Editing the Manuscript, Administration, Supervision, Validation; SAASN: Conceptualization,

Methodology, Programming, Data Curation, Writing and Editing the Original Draft, Project Administration, Supervision.

### Data Availability Statement

The code and information for generating the output are available at https://github.com/mohammad-gh009/Large-Language-Models-vs-Classical-Machine-Learning and https://github.com/Sdamirsa/Tehran_COVID_Cohort. The datasets generated during and/or analyzed during the current study are available from the corresponding author on reasonable request (sdamirsa@ymail.com). We would welcome researchers to build upon our evaluation of LLMs in the context of using structured tabular dataset.

## Additional Information

### Funding

No funding was available related to this work.

### Competing Interests Statement

All the authors declare that they have no financial or nonfinancial conflicts of interest related to this work.

### Abbreviations

LLM: Large Language Model

CML: classical machine learning model

LR: Logistic regression

SVM: Support vector machine

DT: decision tree

KNN: k-nearest neighbor

RF: random forest

XGBoost: Extreme Gradient Boosting

MLP: multilayer perceptron

Zero-shot classification: ZSC

LASSO: least absolute shrinkage and selection operator

SMOTE: synthetic minority oversampling technique

QLoRA: quantized low-lanking adaptation

MICE: Multiple Imputation by Chained Equations

ReLU: rectified linear unit

KBit: knowledge bit

CRP: C-reactive protein

LDH: lactate dehydrogenase

NLP: natural language processing

chain-of-thought (CoT)

This is a supplementary file to "**Large Language Models versus Classical Machine Learning: Performance on COVID-19 Mortality Prediction Using High-Dimensional Tabular Data**" by Mohammadreza Ghaffarzadeh-Esfahani, Mahdi Ghaffarzadeh-Esfahani, Aryan Salahi-Niri, Hossein Toreyhi, Zahra Atf, Amirali Mohsenzadeh-Kermani, Mahshad Sarikhani, Zohreh Tajabadi, Fatemeh Shojaeian, Mohammad Hassan Bagheri, Aydin Feyzi, Mohammadamin Tarighatpayma, Narges Gazmeh, Fateme Heydari, Hossein Afshar, Amirreza Allahgholipour, Farid Alimardani, Ameneh Salehi, Naghmeh Asadimanesh, Mohammad Amin Khalafi, Hadis Shabanipour, Ali Moradi, Sajjad Hossein Zadeh, Omid Yazdani, Romina Esbati, Moozhan Maleki, Danial Samiei Nasr, Amirali Soheili, Hossein Majlesi, Saba Shahsavan, Alireza Soheilipour, Nooshin Goudarzi, Erfan Taherifard, Hamidreza Hatamabadi, Jamil S. Samaan, Thomas Savage, Ankit Sakhuja, Ali Soroush, Girish Nadkarni, Ilad Alavi Darazam, Mohamad Amin Pourhoseingholi, Seyed Amir Ahmad Safavi-Naini

* **Correspondence to:** Seyed Amir Ahmad Safavi-Naini (sdamirsa@ymail.com), Mohamad Amin Pourhoseingholi (aminphg@gmail.com), and Ilad Alavi Darazam (ilad13@yahoo.com)

## Supplementary Section 1. Details and hyperparameters of seven machine learning models

### Supplementary Section 1.1 Logistic Regression

In the logistic regression model employed for this study, L2 regularization (penalty='l2') was applied with a convergence tolerance set to 0.0001, a regularization strength of 1.0, and the use of the limited-memory Broyden–Fletcher–Goldfarb–Shanno algorithm for optimization. The model was configured to fit the intercept term, with the intercept scaling set to 1 and no specific class weights assigned. The logistic regression utilized a maximum of 100 iterations for convergence, automatic detection of the multiclass scenario, 5-fold cross-validation, and a verbosity level set to 0 for minimal output during training. The random state of 42 was set for reproducibility.

### Supplementary Section 1.2 Support Vector Machine

An SVC model was employed in this study with a regularization parameter C set to 1.0, utilizing the radial basis function kernel ('rbf') with a default degree of 3. The gamma parameter was set to 'scale', the coefficient of the kernel function was set to 0.0, and shrinking during optimization was enabled. The model was configured without probability estimates, and the tolerance for convergence was set to 0.001. A cache size of 200 was allocated for optimization (cache_size=200), with no specific class weights assigned (class_weight=None). The model had no specified maximum iteration limit (max_iter=-1) and employed the 'ovr' (one-vs-rest) strategy for decision function shape. Break ties were not considered in the decision function (break_ties=False), 5-fold cross-validation was used, and a random state of 42 was set for reproducibility.

### Supplementary Section 1.3 Decision Tree

A DT classifier was employed in this study with the Gini impurity criterion ('gini'), utilizing the 'best' strategy for splitting nodes. The tree had no specified maximum depth (max_depth=None), and a minimum of 2 samples were required to split an internal node (min_samples_split=2). The minimum number of samples required to be in a leaf node was set to 1 (min_samples_leaf=1), with no specified minimum weight fraction for a leaf node (min_weight_fraction_leaf=0.0). The maximum number of features considered for splitting was not restricted (max_features=None), 5-fold cross-validation was used, and a random state of 42 was set for reproducibility. There were no limitations on the maximum number of leaf nodes (max_leaf_nodes=None), and the minimum impurity decrease for a split was set to 0.0 (min_impurity_decrease=0.0). No specific class weights were assigned (class_weight=None), and the cost complexity pruning alpha (ccp_alpha) parameter was set to 0.0. Additionally, monotonic constraints on features were not specified (monotonic_cst=None).

**Supplementary Section 1.4 K-nearest neighbor**

A KNN classifier was utilized in this study with the number of neighbors set to 5 (n_neighbors=5), employing uniform weights for neighbor contributions ('weights='uniform"). The algorithm used for computing nearest neighbors was automatically determined ('algorithm='auto"), with a leaf size of 30 (leaf_size=30). The Minkowski distance metric with a power parameter of 2 (p=2) was employed ('metric='minkowski"). Additional metric parameters were not specified (metric_params=None), 5-fold cross-validation was used, and parallel processing was not utilized (n_jobs=None).

**Supplementary Section 1.5 Random forest**

An RF classifier was employed in this study with 100 estimators (n_estimators=100), utilizing the Gini impurity criterion ('criterion='gini"). The DT in the RF had no specified maximum depth (max_depth=None), and a minimum of 2 samples was required to split an internal node (min_samples_split=2). The minimum number of samples required to be in a leaf node was set to 1 (min_samples_leaf=1), with no specified minimum weight fraction for a leaf node (min_weight_fraction_leaf=0.0). The square root of the total number of features was considered for splitting at each node ('max_features='sqrt"). The maximum number of leaf nodes was unrestricted (max_leaf_nodes=None), and the minimum impurity decrease for a split was set to 0.0 (min_impurity_decrease=0.0). Bootstrap sampling was enabled (bootstrap=True), out-of-bag scoring was not utilized (oob_score=False), 5-fold cross-validation was used, and parallel processing was not employed (n_jobs=None). A random state of 42 was set for reproducibility during the model-building process (random_state=42). The model verbosity level was set to 0 (verbose=0), and warm starting was disabled (warm_start=False). No specific class weights were assigned (class_weight=None), and the cost complexity pruning alpha (ccp_alpha) parameter was set to 0.0. The maximum number of samples for bootstrapping was not specified (max_samples=None), and monotonic constraints on features were not specified (monotonic_cst=None).

**Supplementary Section 1.6 XGBoost**

The XGBoost classifier model was configured with default hyperparameters, including a base score of 0.5, utilizing a gradient boosting tree-based approach ('booster='gbtree"). The maximum depth of each tree was set to 3, and the learning rate was 0.1. The subsample ratio of training instances was set to 1, indicating the use of all training data. The objective function employed for binary classification was 'binary: logistic', and the number of boosting rounds (n_estimators) was set to 100. The model was configured to use a single CPU core for parallelism (n_jobs=1), 5-fold cross-validation was used, and the random seed number was set to 42 for reproducibility. Default values were applied for parameters such as gamma, min_child_weight, colsample_bytree, reg_alpha, reg_lambda, scale_pos_weight, and verbosity, among others.

**Supplementary Section 1.7 Multilayer Perceptron Neural Network**

A neural network model was employed for classification via the MLP classifier from Scikit-learn. Hyperparameter tuning was conducted through grid search optimization to identify the most effective configuration. The explored hyperparameters included variations in the hidden layer sizes, with options for architectures consisting of a single layer with 100 neurons or two layers with 50 neurons each. The rectified linear unit (ReLU) activation function was chosen, and the Adam solver was employed for optimization. The regularization term (alpha) was set to 0.0001 to control overfitting. The model's training was limited to a maximum of 200 iterations, and a random state of 42 was specified for reproducibility. Early stopping was enabled, with a validation fraction of 0.1 and a criterion of 10 consecutive iterations with no improvement (n_iter_no_change), to halt the training process. The grid search was conducted via 5-fold cross-validation, and the best-performing neural network classifier was identified as the one with the optimal combination of hyperparameters. The resulting best classifier, determined through the grid search, was then further evaluated on the training data, and its performance metrics were used for subsequent analysis and interpretation.

## Supplementary Section 2. Details of fine-tuning the LLM with the QLoRA

QLoRA is a novel and efficient fine-tuning approach designed to reduce memory usage significantly. QLoRA introduces innovations such as a new 4-bit data type, double quantization to reduce the memory footprint, and paged optimizers to manage memory spikes. The approach demonstrates superior performance across various instruction datasets, model types (LLaMA, T5), and scales (e.g., 33B and 65B parameter models), showing state-of-the-art results through fine-tuning on a small high-quality dataset (1).

We enabled 4-bit loading, employed double quantization, utilized a quantization type of "nf4," and specified the computed data type as torch.bfloat16 for our model. We then created a tokenizer via the pretrained model and generated a causal language model, leveraging quantization configurations from the specified BitsAndBytesConfig ("bnb_config"). We enabled gradient checkpointing for our model and prepared it for knowledge bit (KBit) training via the prepare_model_for_kbit_training function from the parameter-efficient fine-tuning (PEFT) library.

We utilized the PEFT library to create a LoraConfig object with specified parameters, including a 16-layer model with Lora attention, targeted projection modules, a dropout rate of 0.1, no bias, and a task type of 'CAUSAL_LM'. We subsequently generated a PEFT model on the basis of this configuration.

We utilized the transformer library to establish a training pipeline for our language model. The pipeline involved initializing a trainer with our specified model, training dataset, and training arguments. The training configuration entailed setting the output directory to "logs" and conducting training for 4 epochs. We utilized a per-device batch size of 1, with gradient accumulation steps of 4. The chosen optimizer was "paged_adamw_32bit", with a learning rate of 2e-4 and a weight decay of 0.001. We employed mixed-precision training with fp16 while disabling boff16. To control the maximum gradient norm, we set it to 0.3. The maximum number of training steps was not limited (-1), and we incorporated a warm-up ratio of 0.03 and enabled grouping by sequence length. The learning rate scheduler employed was "cosine". The training progress was reported to TensorBoard, and evaluation was performed at the end of each epoch. We disabled saving checkpoints during training by setting save_steps to 0. Additionally, we set logging_steps to 25 to log training progress at regular intervals. We specifically explored QLORA, combined with the bits and bytes library, to enhance language models while requiring fewer resources.

## Supplementary Tables

### Supplementary Table S1. Description of the raw prompts and improved prompts used during the experiments.

| Experiment | Raw Prompt | Improved Prompts | Improved prompt after 5 attempts |
|---|---|---|---|
| All LLMs including: Zero-shot classification including Mixtral-8x7B-Instruct-v0.1, Llama-3-70B, Mistral-7B-Instruct, Llama-3-8B, gpt-4o, gpt-3.5-turbo, gpt-4-turbo(g), gpt-4 | "Does the patient survive or die based on the provided medical history? patient history is: "+["patient medical history"] | You're tasked with analyzing the present symptoms, past medical history, aboratory data, age, and gender of COVID-19 patients to determine their outcome, which is enclosed in square brackets. Your goal is to predict whether the patient will "survive" or "die" based on this information. | You're tasked with analyzing the present symptoms, past medical history, laboratory data, age, and gender of COVID-19 patients to determine their outcome, which is enclosed in square brackets. Your goal is to predict whether the patient will "survive" or "die" based on this information. Predict patient mortality in JUST ONE word and DO NOT answer vaguely. |
| Fine-tuned LLM (mistral-7b-instruct-v0.2) | (This prompt was not used in the fine-tuning task.) | You're tasked with analyzing the present symptoms, past medical history, laboratory data, age, and gender of COVID-19 patients to determine their outcome, which is enclosed in square brackets. Your goal is to predict whether the patient will "survive" or "die" based on this information. | (This prompt was not used in the fine-tuning task.) |

**Supplementary Table S2. Environments, model settings and model parameters for each LLM**

| Model name in paper | Model names | Environment and model setting | Model parameter | Date/Source of use |
|---|---|---|---|---|
| Mixtral-8x7b | Mixtral-8x7b-Instruct-v0.1 | Poe Web Interface | 46.7 b | April 2024 |
| Llama3-70b | Llama-3-70b | Poe Web Interface | 70B | May 2024 |
| Mistral-7b | Mistral-7b-Instruct | Poe Web Interface | 7 B | April 2024 |
| Llama3-8b | Llama-3-8b | Poe Web Interface | 8 B | May 2024 |
| GPT4o | gpt-4o-2024-05-13 | OpenAI API: Temperature 1; max_tokens: 1024; seed: 123 | 175 B | June 2024 |
| GPT3.5 | gpt-3.5-turbo-0125 | OpenAI API: Temperature 1; max_tokens: 1024; seed: 123 | 20 B | April 2024 |
| GPT4T | gpt-4-turbo-2024-04-09 | OpenAI API: Temperature 1; max_tokens: 1024; seed: 123 | 175 B | June 2024 |
| GPT4 | gpt-4-0613 | OpenAI API: Temperature 1; max_tokens: 1024; seed: 123 | 175 B | June 2024 |
| BERT (2) | bert-base-uncased | Transformers library (Python) | 110 M | huggingface.co/google-bert/bert-base-uncased |
| ClinicalBERT[a] (3) | - | Transformers library (Python) | 110 M | huggingface.co/medicalai/ClinicalBERT |

Footnote: a: The ClinicalBERT model was developed using a large multicenter dataset comprising 1.2 billion words covering a wide range of diseases. This base language model was further fine-tuned via electronic health records (EHRs) from over 3 million patient records to enhance its performance in clinical settings.

**Supplementary Table S3. Comparison of XGBoost performance with and without imputation.**

| | Approach | Accuracy | Precision | Recall | Specificity | F1 | AUC |
|---|---|---|---|---|---|---|---|
| **Internal validation** | | | | | | | |
| | Xgboost without imputation | 0.84 | 0.87 | 0.94 | 0.43 | 0.91 | 0.68 |
| | Xgboost with imputation | 0.87 | 0.88 | 0.84 | 0.89 | 0.86 | 0.95 |
| **External validation** | | | | | | | |
| | Xgboost without imputations | 0.82 | 0.83 | 0.96 | 0.25 | 0.89 | 0.60 |
| | Xgboost with imputation | 0.82 | 0.85 | 0.78 | 0.86 | 0.82 | 0.92 |

**Supplementary Table S4. Comparison of CML's performance with and without SMOTE, compared to our pipiline with SMOTE**

| | Model | SMOTE | Accuracy | Precision | Recall | Specificity | F1 | AUC |
|---|---|---|---|---|---|---|---|---|
| **Internal Validation** | | | | | | | | |
| | LR | No | 0.81 | 0.86 | 0.9 | 0.4 | 0.88 | 0.65 |
| | | Yes | 0.74 | 0.74 | 0.75 | 0.74 | 0.74 | 0.82 |
| | SVM | No | 0.84 | 0.86 | 0.96 | 0.33 | 0.9 | 0.64 |
| | | Yes | 0.76 | 0.73 | 0.81 | 0.71 | 0.77 | 0.85 |
| | DT | No | 0.75 | 0.86 | 0.83 | 0.43 | 0.85 | 0.63 |
| | | Yes | 0.76 | 0.76 | 0.76 | 0.76 | 0.76 | 0.76 |
| | KNN | No | 0.82 | 0.84 | 0.96 | 0.22 | 0.89 | 0.59 |
| | | Yes | 0.69 | 0.71 | 0.66 | 0.72 | 0.68 | 0.74 |
| | RF | No | 0.84 | 0.87 | 0.94 | 0.42 | 0.9 | 0.68 |
| | | Yes | 0.86 | 0.84 | **0.89** | 0.83 | **0.87** | 0.94 |
| | Xgboost | No | 0.82 | 0.88 | 0.9 | 0.48 | 0.89 | 0.69 |
| | | Yes | **0.87** | **0.88** | 0.84 | **0.89** | 0.86 | **0.95** |
| | MLP | No | 0.83 | 0.88 | 0.91 | 0.5 | 0.9 | 0.7 |
| | | Yes | 0.76 | 0.73 | 0.81 | 0.70 | 0.77 | 0.83 |
| **External Validation** | | | | | | | | |
| | LR | No | 0.81 | 0.84 | 0.95 | 0.29 | 0.89 | 0.65 |
| | | Yes | 0.73 | 0.73 | 0.74 | 0.72 | 0.73 | 0.80 |
| | SVM | No | 0.82 | 0.84 | 0.95 | 0.29 | 0.89 | 0.64 |
| | | Yes | 0.74 | 0.73 | 0.75 | 0.72 | 0.74 | 0.81 |
| | DT | No | 0.72 | 0.85 | 0.77 | 0.48 | 0.81 | 0.63 |
| | | Yes | 0.73 | 0.72 | 0.73 | 0.72 | 0.73 | 0.72 |
| | KNN | No | 0.79 | 0.83 | 0.94 | 0.24 | 0.88 | 0.59 |
| | | Yes | 0.68 | 0.71 | 0.62 | 0.75 | 0.66 | 0.73 |
| | RF | No | 0.81 | 0.85 | 0.92 | 0.34 | 0.88 | 0.68 |
| | | Yes | **0.82** | 0.79 | **0.86** | 0.77 | **0.83** | 0.91 |
| | Xgboost | No | 0.78 | 0.86 | 0.87 | 0.44 | 0.86 | 0.69 |
| | | Yes | **0.82** | **0.85** | 0.78 | **0.86** | 0.82 | **0.92** |
| | MLP | No | 0.79 | 0.86 | 0.88 | 0.42 | 0.87 | 0.7 |
| | | Yes | 0.71 | 0.69 | 0.76 | 0.67 | 0.72 | 0.79 |

## Supplementary Figures

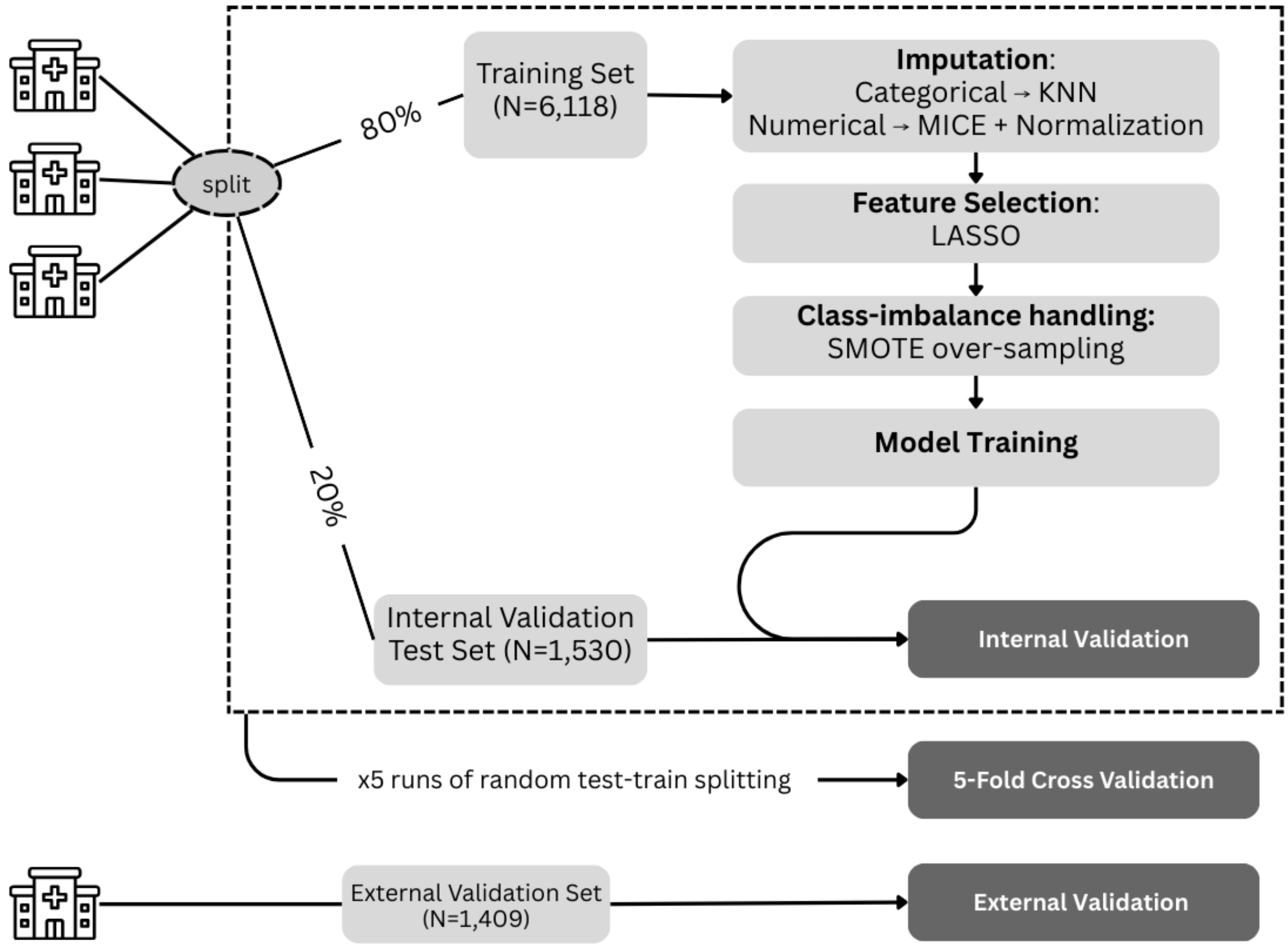

**Supplementary Figure S1. The complete data flow from raw data preprocessing through training-test splitting, LASSO-based feature selection on the training set, and subsequent model training and evaluation.**

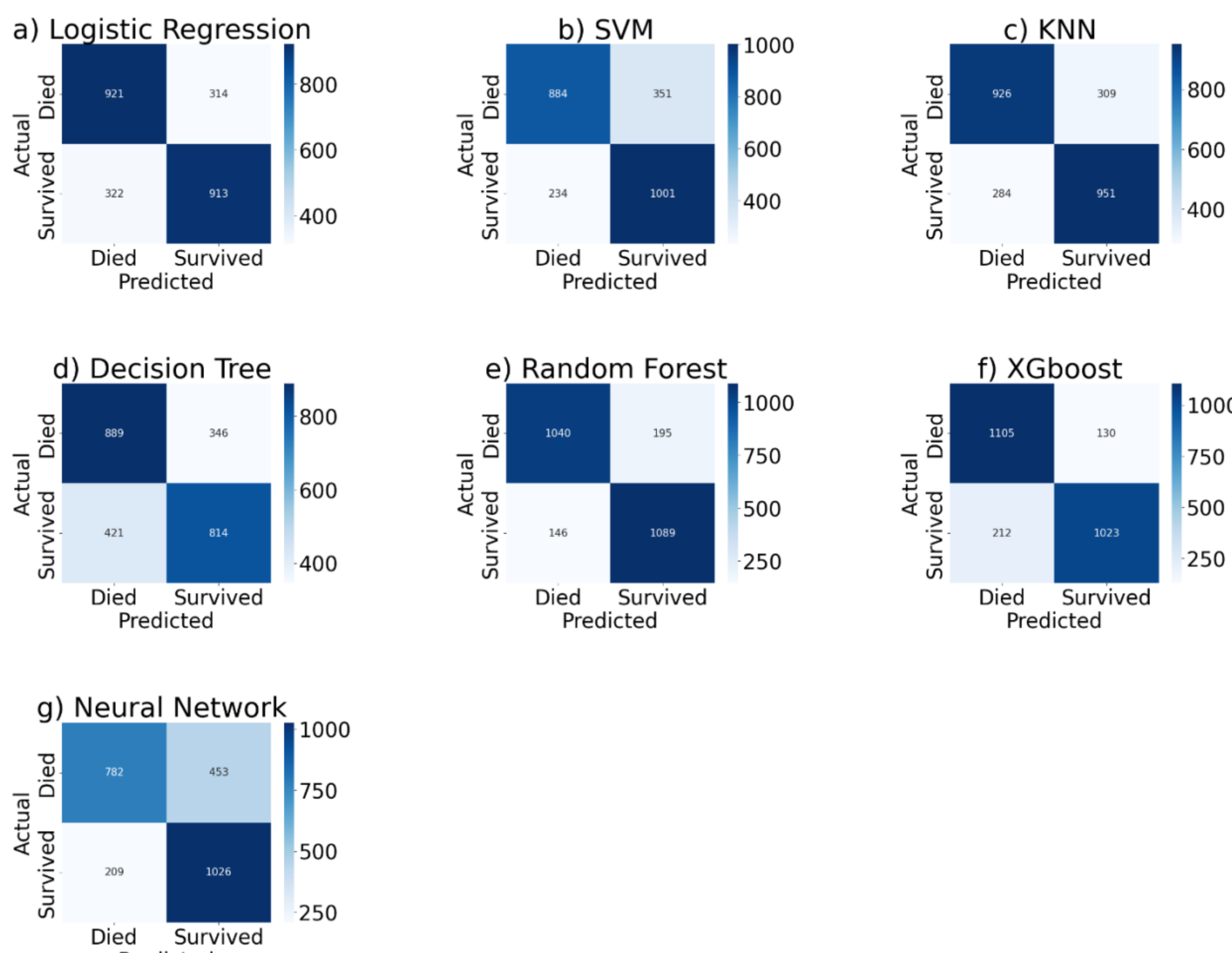

**Supplementary Figure S2. Confusion matrices for classical machine learning models (CMLs) trained on the internal validation test set, including logistic regression (a), support vector machine (b), K-nearest neighbor (c), decision tree (d), random forest (e), XGBoost (f), and multilayer perceptron (MLP)**

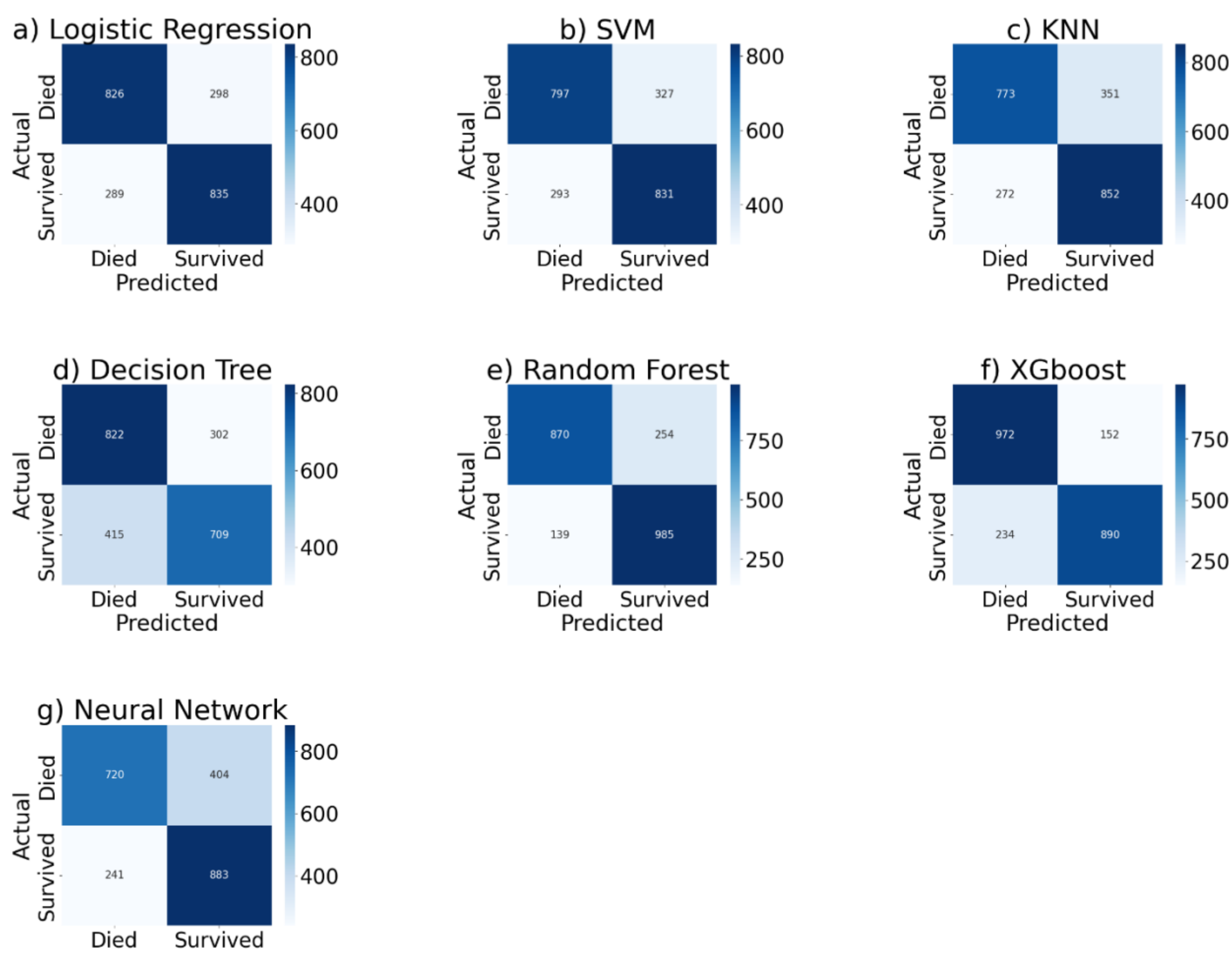

**Supplementary Figure S3. Confusion matrices for seven classical machine learning models tested on the external validation set: logistic regression (a), support vector machine (b), K-nearest neighbor (c), decision tree (d), random forest (e), XGBoost (f), and multilayer perceptron (MLP) (g)**

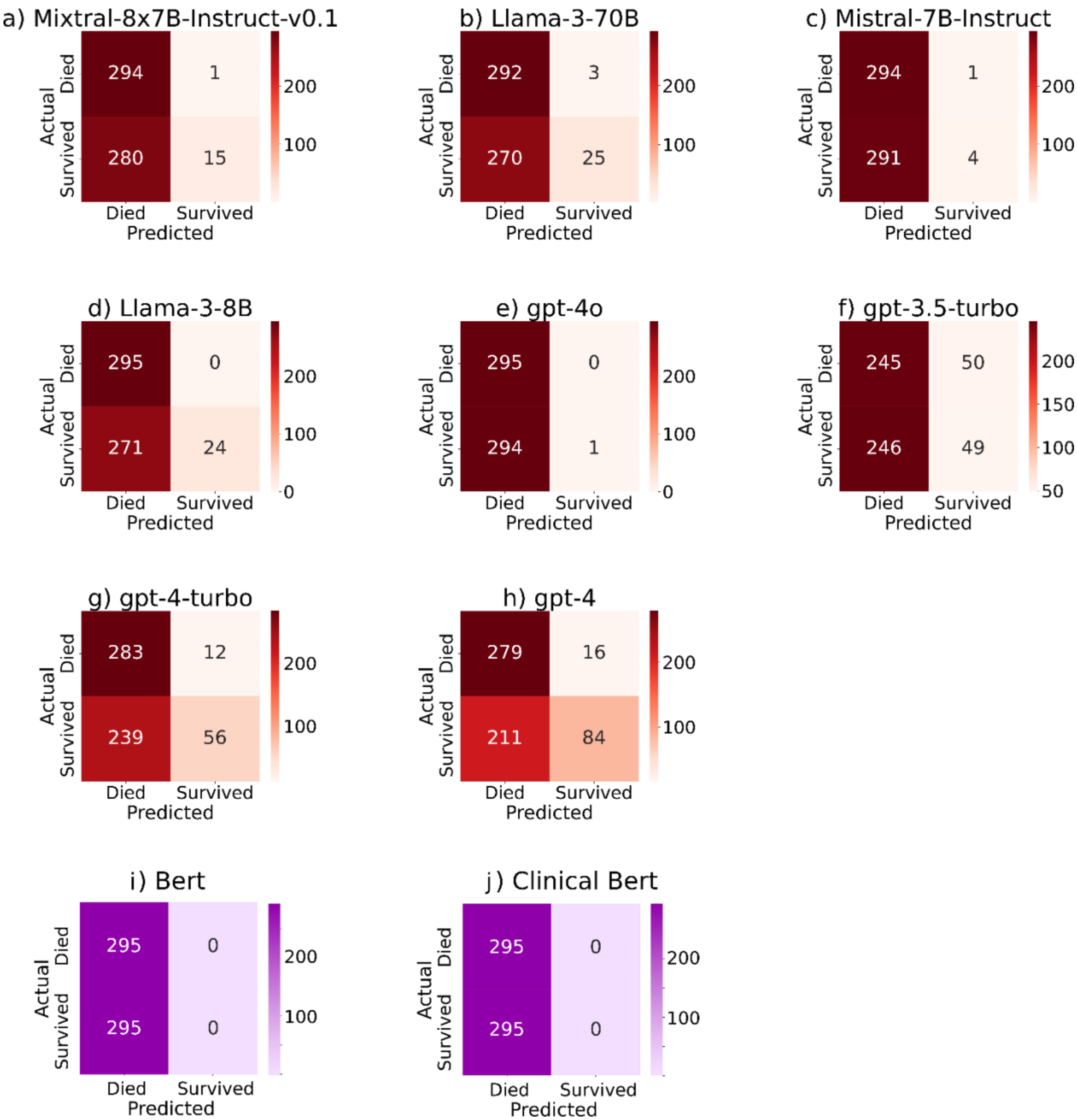

**Supplementary Figure S4. Confusion matrices for LLM zero-shot tasks, including Mixtral-8x7B-Instruct-v0.1 (a), Llama-3-70B (b), Mistral-7B-Instruct (c), Llama-3-8B (d), gpt-4° (e), gpt-3.5-turbo (f), gpt-4-turbo (g), and gpt-4 (h)**

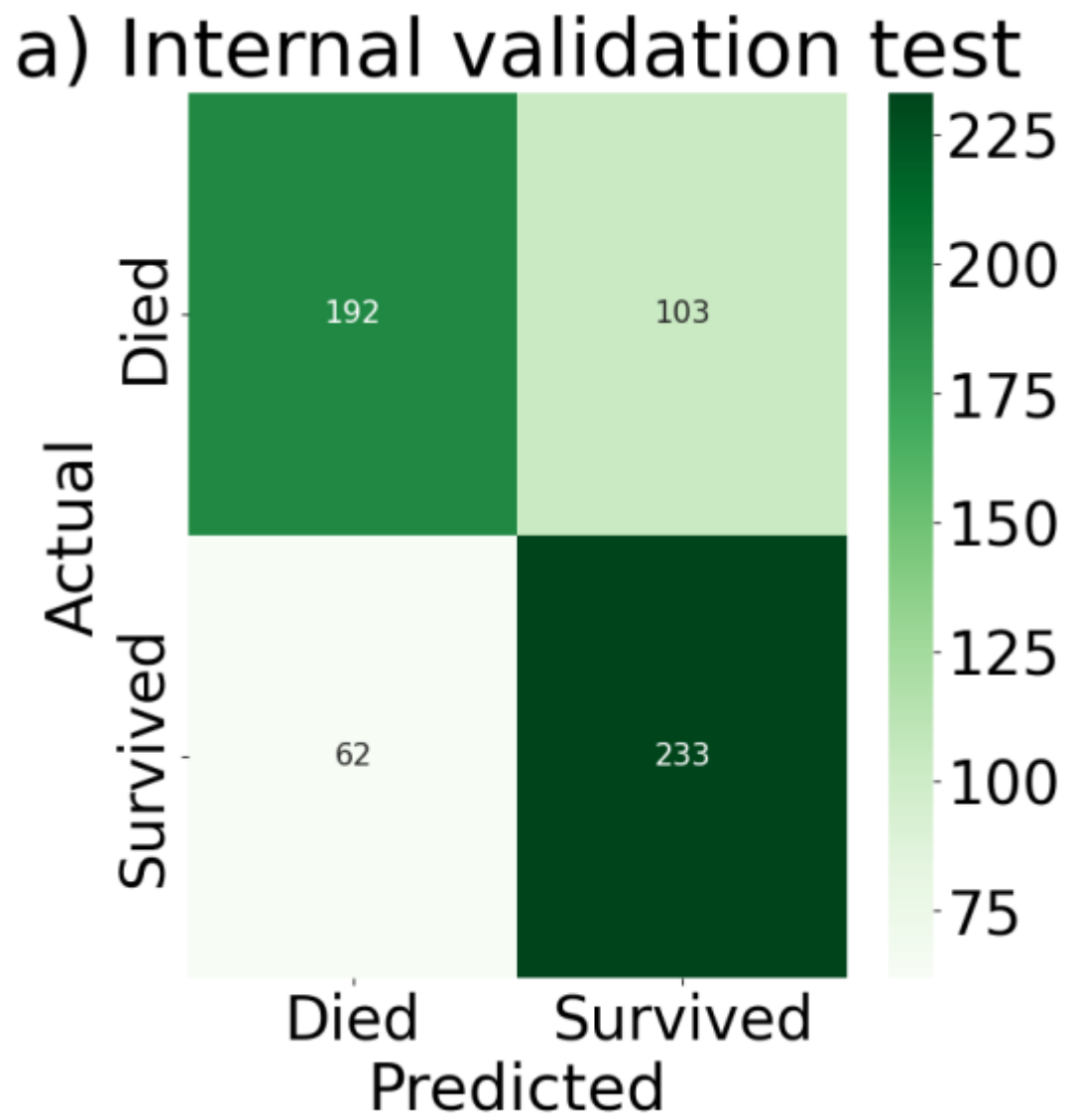

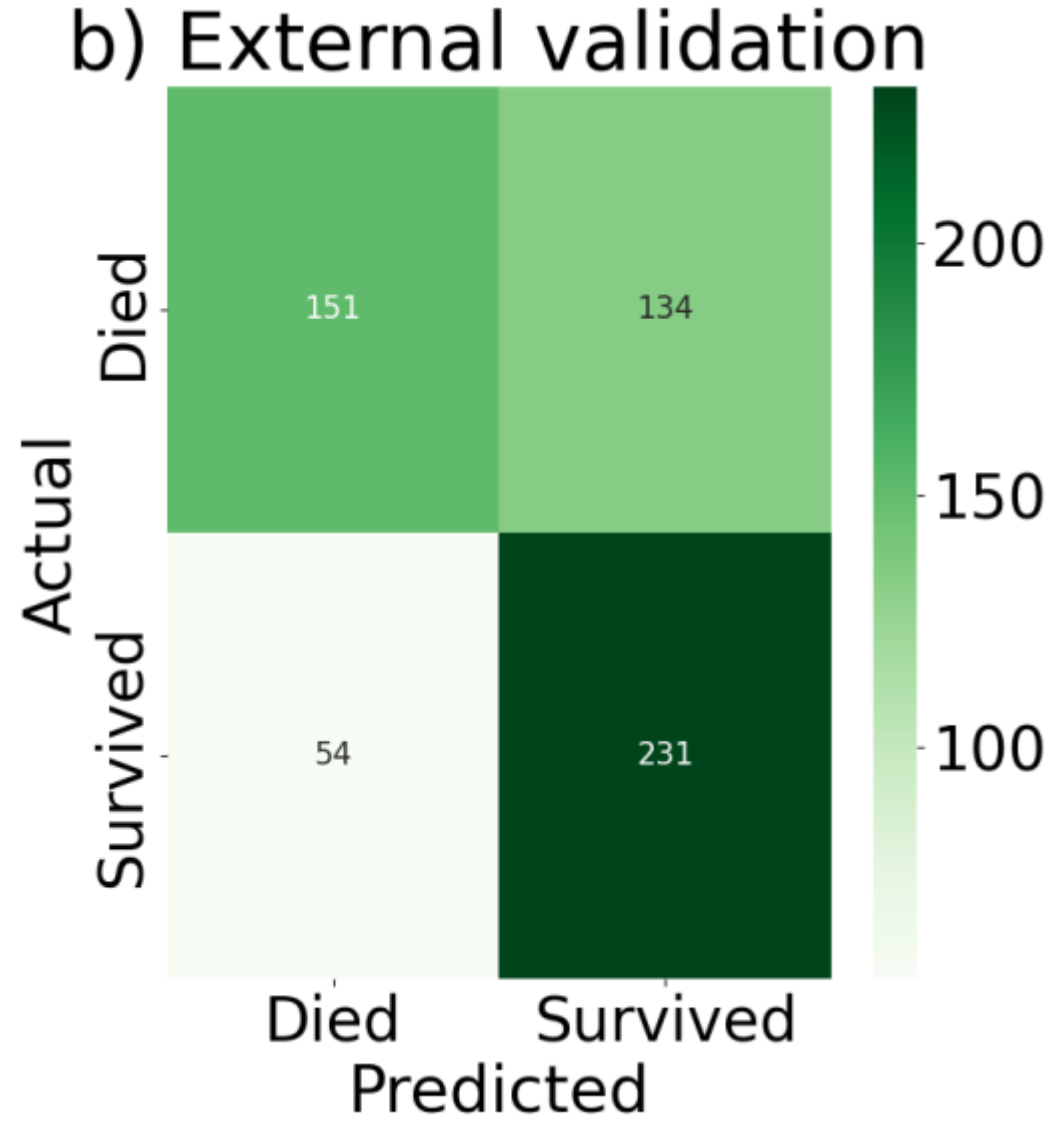

**Supplementary Figure S5. Confusion matrices for fine-tuned mistral-7b-instruct-v0.2 on the internal validation test set (a) and external validation set (b).**

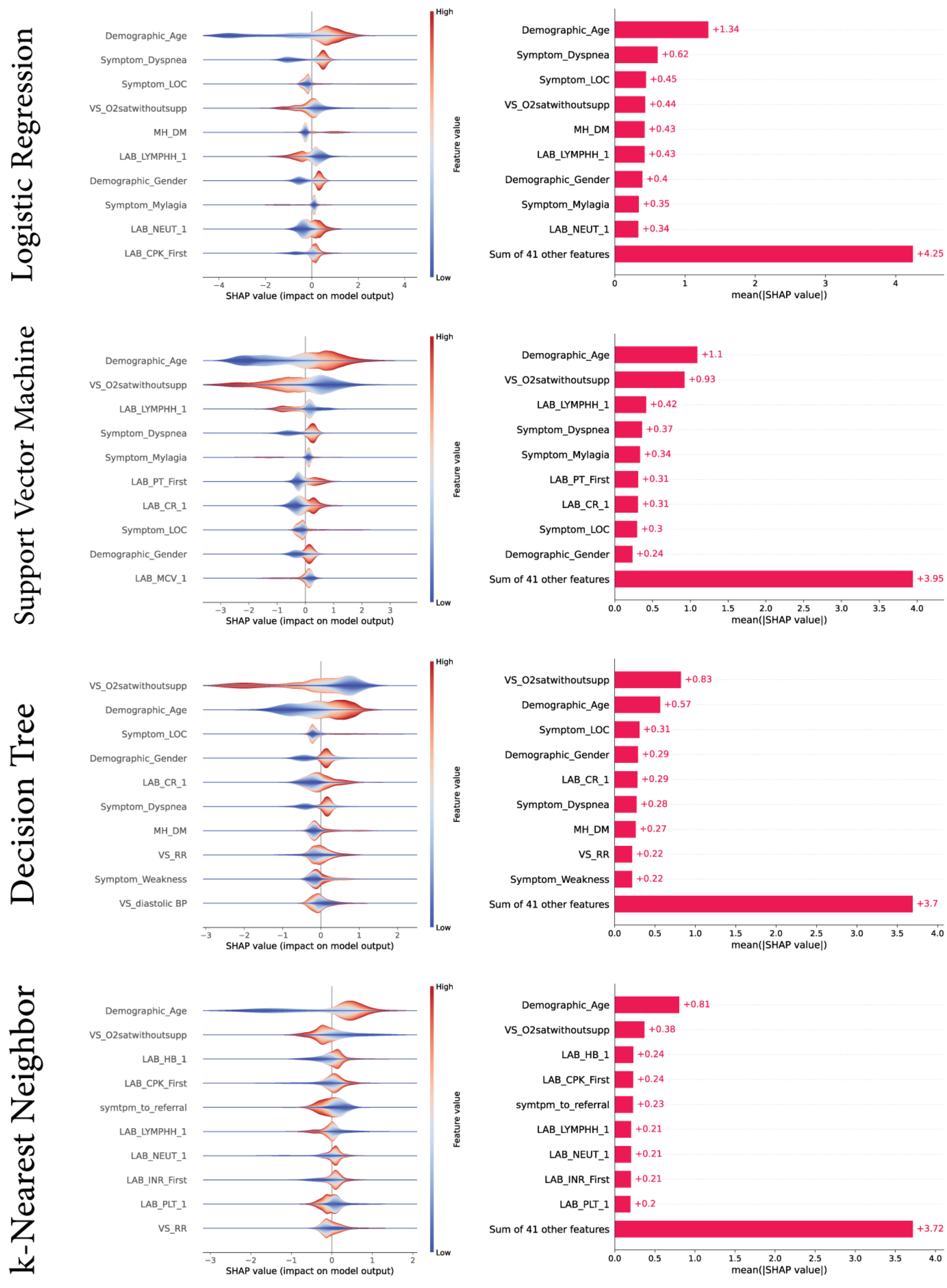

**Supplementary Figure S6. Features with granular (left column) and global (right column) impacts on predictions among CMLs**

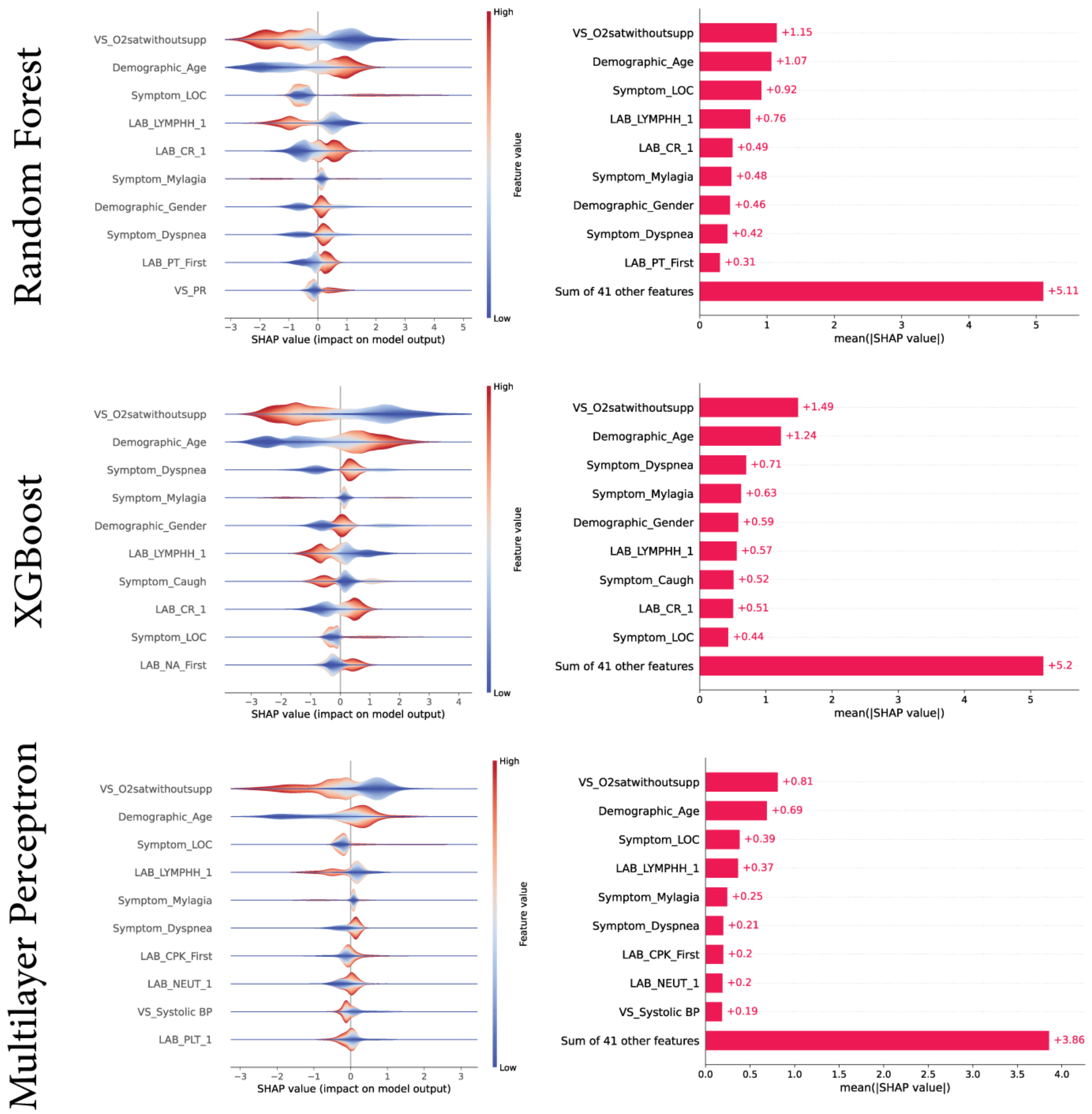

**Supplementary Figure S6 (continued). Features with granular (left column) and global (right column) impacts on predictions among CMLs**

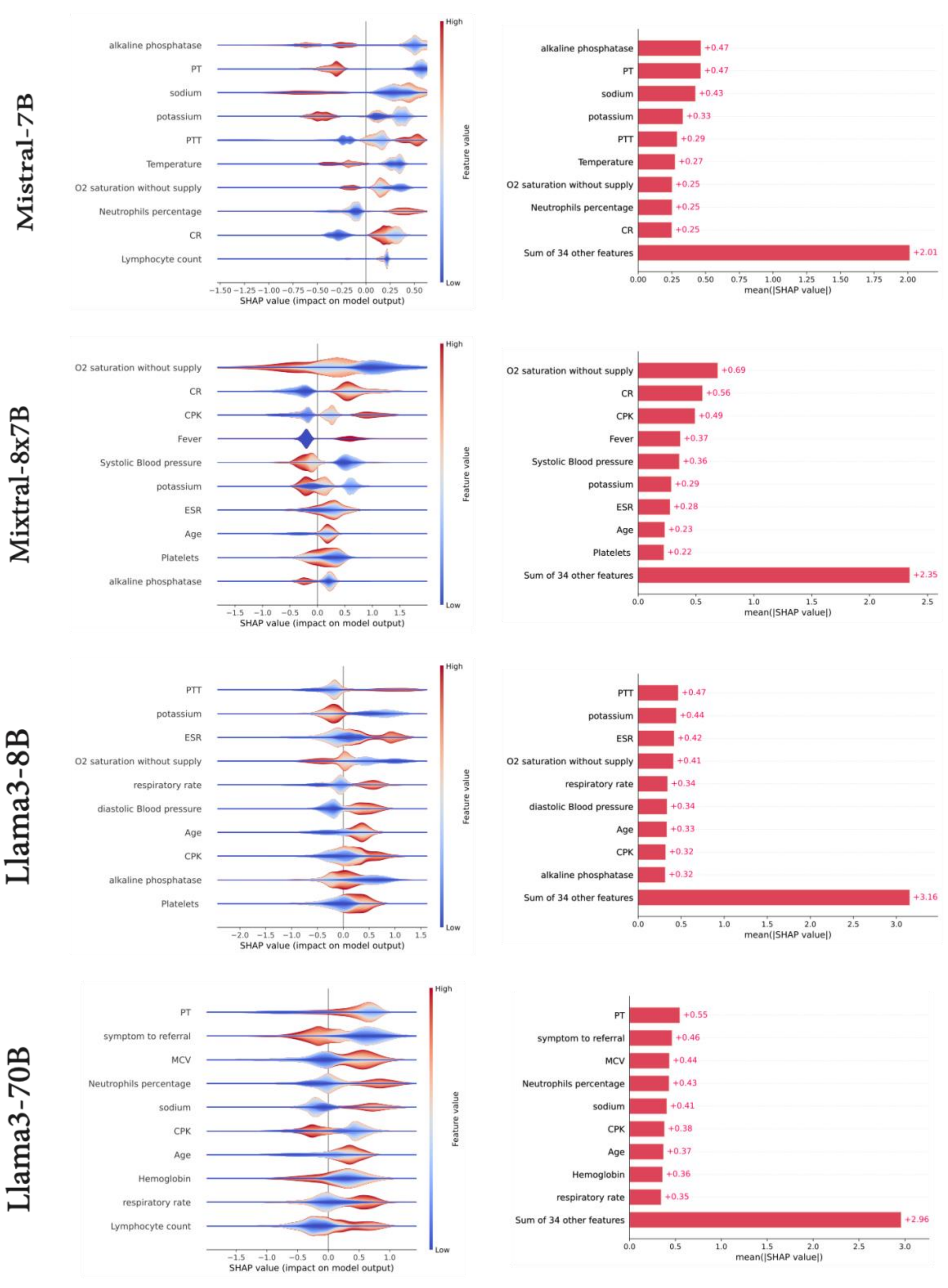

**Supplementary Figure S7. Features with granular (left column) and global (right column) impacts on predictions among LLMs**

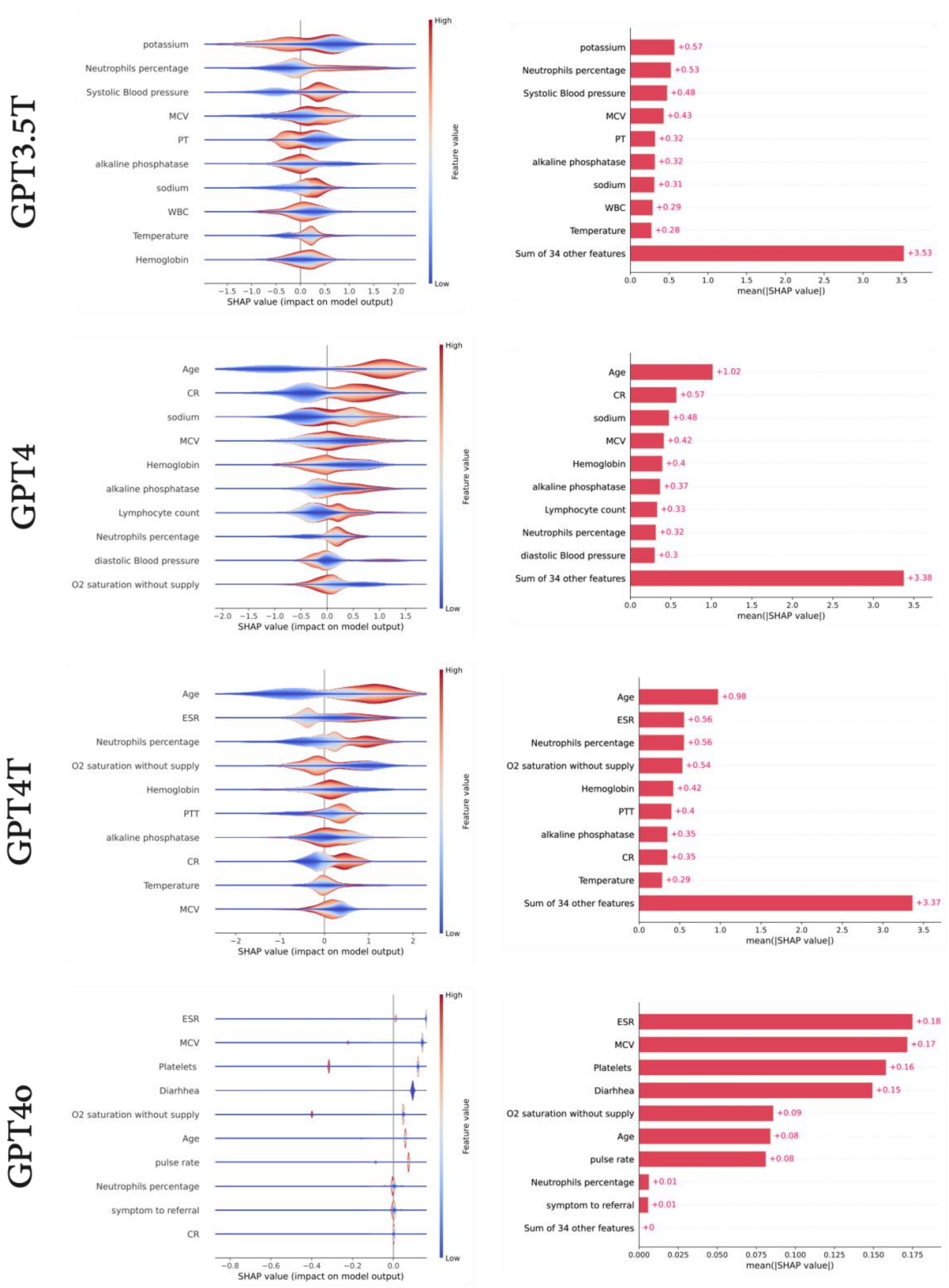

**Supplementary Figure S7 (continued). Features with granular (left column) and global (right column) impacts on predictions among LLMs.**